\pdfoutput=1

\documentclass[11pt]{article}

\usepackage[preprint]{acl}

\usepackage{times}
\usepackage{latexsym}

\usepackage{tabularx}
\usepackage{multirow}
\usepackage{multicol}
\usepackage{xurl}  
\usepackage{hyperref}
\usepackage{booktabs}  

\usepackage[T1]{fontenc}

\usepackage[utf8]{inputenc}

\usepackage[x11names,dvipsnames]{xcolor}
\usepackage{microtype}

\usepackage{inconsolata}

\usepackage{graphicx}

\usepackage{tabularx}
\usepackage{amsmath}

\title{Computational Analysis of Character Development\\ in Holocaust Testimonies}

\author{Esther Shizgal$^1$\quad Eitan Wagner$^1$\quad Renana Keydar$^2$\quad Omri Abend$^1$ \\
         $^1$ Department of Computer Science \quad
         $^2$ Faculty of Law and Digital Humanities\\
         Hebrew University of Jerusalem\\ \texttt{\{first\_name\}.\{last\_name\}@mail.huji.ac.il}}

\usepackage{graphicx}
\usepackage{subcaption}
\usepackage{caption}
\usepackage{todonotes}

\begin{document}
\maketitle
\begin{abstract}
This work presents a computational approach to analyze character development along the narrative timeline. 
The analysis characterizes changes in the protagonist's views and behavior and the interplay between them. 
We consider transcripts of Holocaust survivor testimonies as a test case, each telling the story of an individual in first-person terms.
We focus on the survivor’s religious trajectory, examining the evolution of their disposition toward religious belief and practice as it is reflected in the testimony. Clustering the resulting trajectories in the dataset, we identify common sequences in the data. 
Our findings highlight multiple common structures of religiosity across the narratives: in terms of belief, a constant disposition is common, while for practice, most present an oscillating structure, serving as valuable material for historical and sociological research. This work demonstrates the potential of natural language processing for analyzing character evolution through thematic trajectories in narratives.
\end{abstract}

\section{Introduction}
\label{intro}

Characters in narratives are shaped by the continuous interplay of internal and external events they experience, each influencing their thoughts, feelings, and behaviors. As narratives progress, these events unfold along various dimensions. Some are common to essentially all narratives, such as changes in location, while others are specific to particular genres \cite{eder2010understanding}. Personal narratives, told in the first person, offer insights into the inner-life development of the protagonist \cite{lifestories}.

Within the domain of Holocaust testimonies, as people experience unimaginable atrocities, the change in religious dispositions plays a central role \cite{Brenner1980TheFA}. \citet{schweid1988faith} discusses the post-Holocaust crisis of theodicy in Jewish and Christian religious thought, necessitating substantial changes in religious content and norms. Specifically, the question of how religious faith evolves under extreme circumstances remains actively debated in Holocaust studies, with scholars like \citet{berkovits1973faith} stressing that the majority of survivors maintained their faith, and \citet{Leo2021} discussing the role of faith as a stabilizing force, arguing that most individuals do not change their religious beliefs following traumatic experiences. In contrast, \citet{benezra} describe common processes of change in both directions. This ongoing argument makes Holocaust testimonies a suitable test case for religious evolution analysis, offering further insight by analyzing a large corpus of testimonies through a holistic computational approach that could extend to other dimensions of character development. 

Holocaust survivor testimonies are indispensable for both scholarship and collective memory. Applying NLP to historical texts, and in particular to these testimonies, has been advocated \citep{artstein-etal-2016-new, wagner-etal-2022-topical}, offering the potential to extract valuable insights from the vast number of testimonies, instead of relying on small-scale, manual studies \citep{ToddPresner, garcia2023if}. Following this approach, we utilize NLP technology to examine religious trajectories across an entire corpus of testimonies, instead of focusing on individual cases.
Leveraging off-the-shelf LLMs for inferring the character arc of the protagonist, we explore the progression of their religious trajectory throughout the narrative. We conduct a dual perspective analysis that captures the protagonist's descriptions of their beliefs and religious practices.

\begin{figure*}[t!]
    \centering
    \includegraphics[width=0.9\linewidth]{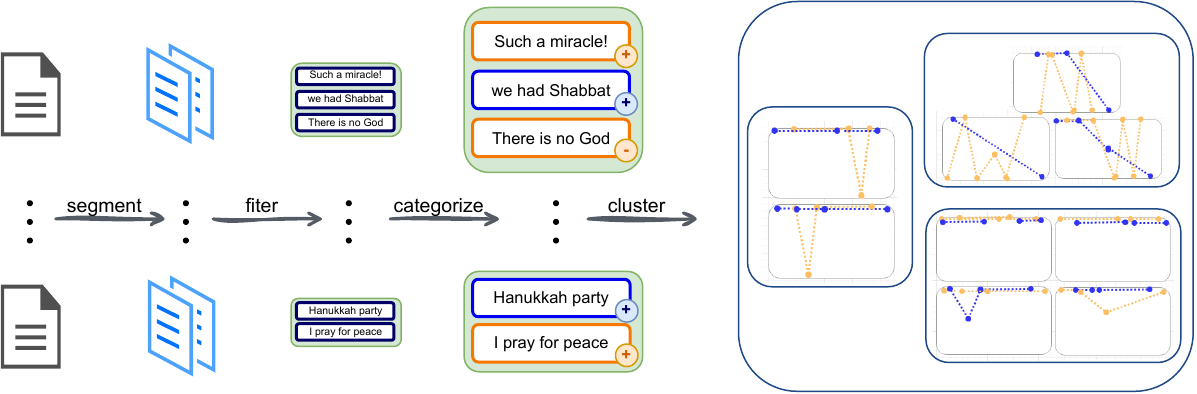}
    \caption{Our pipeline for identifying the religious trajectories from a set of Holocaust survivor testimonies: (1) \textbf{Segmentation:} segment the testimonies by question-answer pairs; 
    (2) \textbf{Filtration:} train and run a classifier to filter all segments containing religious content; 
    (3) \textbf{Determining Valence:} use LLMs to identify the protagonist's valence of religious practice and/or beliefs in a given segment;
    (4) \textbf{Schematization:} cluster the resulting trajectories to identify common patterns of evolution of religiosity in the testimony dataset.}
    \label{fig:pipeline-palceholder}
\end{figure*}

As people change and adapt, their disposition towards religion and level of observance constantly evolve. This is particularly relevant when addressing Holocaust testimonies, which follow survivors’ experiences in extreme conditions. We expect to reveal different patterns of change in religiosity throughout the testimonies, each telling the story of an individual and covering a wide range of periods and locations.  

We define a {\it religious trajectory} as a series of mentions of religiosity throughout a given storyline of an individual's life course. Separate trajectories are held for religious practices and beliefs, assuming they each describe religiosity from a unique perspective. This distinction is discussed in the sociology of religion and is reflected by incongruities among religious observers. \citet{Chaves2010SSSRPA} illustrates this: \textit{``Ideas and practices exist as bits and pieces that come and go as situations change, producing many inconsistencies... Religious ideas, values, and practices generally are not congruent.''}

For extracting the trajectories, we identify the segments within the text that constitute the trajectory, and assess the protagonist's religious valence and intensity in each segment. We then cluster the resulting trajectories, identifying common sequences (\autoref{fig:pipeline-palceholder}). Our research contributes to a deeper understanding of character evolution within narratives by highlighting the potential of machine learning techniques to study thematic trajectories.

\section{Related Work}

\paragraph{Narrative Analysis.} Recent NLP research has underscored the importance of narrative analysis in the study of human behavior and belief systems. While these studies often center around event and location schemas \cite{wagner2025unsupervised}, character personalities, narrator intentions \cite{piper-etal-2021-narrative, zhu-etal-2023-nlp}, or narrative structure \cite{wagner-etal-2022-topical}, there remains a gap in tracking the evolution of characters, particularly their development. This gap is crucial, as understanding character evolution can provide deeper insight into how narratives convey complex human experiences.

Another aspect studied is character attribute inference, analyzing personas or archetypes, relationships, and emotional trajectories \cite{chaturvedi-etal-2017-story}. An example of studying thematic trajectories of characters is \citep{brahman-etal-2021-characters-tell}, which models the emotional trajectory of the protagonist by generating and modeling stories that follow various emotional arcs. 

This study builds upon these foundations by analyzing character development, focusing on the protagonist's described religious practices and their expressed religious belief systems. 

\paragraph{Religious Trajectories.} 
Our analysis relies on the ability of LLMs to recognize various valences of religious activity and beliefs, including sacred emotions associated with spiritual or religious practices. \citet{plaza-del-arco-etal-2024-divine} explore how religion is represented in LLMs, showing that these models can attribute emotions tied to religion, and display some awareness of sacred emotions, although in an inconsistent and biased manner. 

In the social sciences, much work has examined religious trajectories with \citet{IngersollDayton2002ReligiousTA} analyzing patterns of change and stability in religiosity over the life course, and uncovering four distinct patterns of religious trajectories: stable, increasing, decreasing, and curvilinear. Similarly, \citet{McCullough2005TheVO} study religious trajectories among immigrant populations. 
More recently, \citet{Exline2020PullingAF} explore the religious and spiritual struggles of college students distancing from religion, including conflicts and
intellectual questions about God's existence.

In the context of the Holocaust, \citet{defectiveCovenant} examines religious transformation among Jewish survivors of the Holocaust, focusing on testimonies that describe a shift from being connected with religious traditions to abandoning belief in God completely. This work discusses how survivor testimonial information can serve as valuable sources regarding faith after the Holocaust.

We leverage recent advances in NLP to analyze religious trajectories on a broader scale using a large dataset of personal testimonies. Rather than focusing on individual cases, our approach aims to uncover common patterns and sequences across many narratives, providing insights into how religious beliefs and practices evolve throughout the life course and under trauma, within this context.

\paragraph{Trajectory Clustering.}
\citet{Alqahtani2021DeepTC} held a thorough review of time-series data analysis, focusing on deep time-series clustering (DTSC), proposing a clustering approach using deep convolutional auto-encoders (DCAE). \citet{Chang2023TrajectorySM} discuss methods for this task, including traditional metrics for trajectories, and deep learning approaches that embed trajectories for similarity measurement. However, these methods typically require labeled data for supervised learning. 

\section{Task Definition}

Given a segmented testimony $(x_1, x_2, \ldots, x_n)$, and a set of labels for practice and belief content $\mathcal{B}$ and $\mathcal{P}$, our framework outputs a sequence representing its religious trajectory $(\langle b_1, p_1\rangle, \langle b_2, p_2\rangle, \ldots, \langle b_n, p_n\rangle)$, where $b_j \in \mathcal{B} \cup \{\texttt{None}\}$ and  $p_j \in \mathcal{P} \cup \{\texttt{None}\}$. The \texttt{None} label is used for segments that do not contain belief or practice content.

\section{Data and Annotation}

We received 1000 Holocaust survivor testimony transcripts archived by the Shoah Foundation\footnote{\url{https://sfi.usc.edu/}} (SF), originally recorded as interviews on video in English. The significance of this resource in the study and remembrance of the Holocaust cannot be overstated. As the number of surviving witnesses diminishes, there is an urgent need to find new approaches to engage with the extensive collection of testimonies, with computational methods offering interdisciplinary pathways to trace patterns of belief, resilience, identity, and trauma \citep{wagner2025unsupervised, toth2022studying}.

Due to the lack of a structured format, our first step is to segment the data using natural divisions created by question-answer pairs followed by merging segments with fewer than 10 words and dividing those exceeding 100 words.
The method aims to create segments that contain just enough information to determine its relevance to the religious trajectory. This method yields a dataset with the majority of the segments being between 50-100 words and reasonable paragraph breaks, mostly keeping separate ideas distinct. However, some samples may contain multiple topics or conflicting religious signals. For example: \textit{``Saturday I would go, and I was encouraged to go to the synagogue. And Sunday, I was encouraged to go to church.''} 

Our analysis is grounded in comparing religious trajectories along their narrative timelines, which raises the question of how the chronological order of events and the narrative sequence align. Examining this relationship shows they are indeed closely aligned (see \autoref{chronological_time}).

\subsection{Annotations}
After segmenting the data, we randomly sample and annotate approximately 4000 segments. 
The first annotation assignment is for capturing descriptions of Jewish religious content.
This is designed as a binary classification task -- the annotators are directed to classify all segments that describe any Jewish religious practices or beliefs of the protagonist or indicate their absence. Complete guidelines are provided in \autoref{sec:appendix_annotations}.

This complex annotation task involves knowledge of history and the Jewish religion. Many segments can be ambiguous, rely on context, or require an understanding of the context. Therefore, there may be a lack of consensus among the annotators.
All these factors contribute to the complexity of the task, which requires careful attention.

The annotation process was carried out by three annotators. We keep an overlap of 833 samples annotated by at least two annotators, for measuring inter-annotator agreement (IAA). To evaluate the IAA, we use Krippendorff's alpha \cite{krippendorff2011computing} on the pairwise annotations, obtaining an average score of 0.74, indicating a substantial level of agreement. \autoref{detailed_breakdown} presents the label distribution of the overlap set of annotations.

In case of disagreement within the overlap: if only two annotators reviewed the sample it was discarded from the dataset. Otherwise, the majority label was assigned. Due to the challenge of collecting annotations, the rest of the samples were annotated by a single annotator. Additionally, throughout the annotation process, we occasionally review the annotations and ask the annotators to revise their work when necessary.

The second annotation task is multi-label and consists of identifying the class of each sample and
determining the speaker’s valence toward it. Each segment belonging to the trajectory is independently categorized by its content – practice, belief, or both (see \S \ref{intro}). The segments are further categorized by their valence toward the specific practice or belief
they describe.
Specifically, we map each sample to one or more of the classes or mark it as falsely recognized religious content.

\begin{figure}[t]
  \includegraphics[width=\columnwidth]{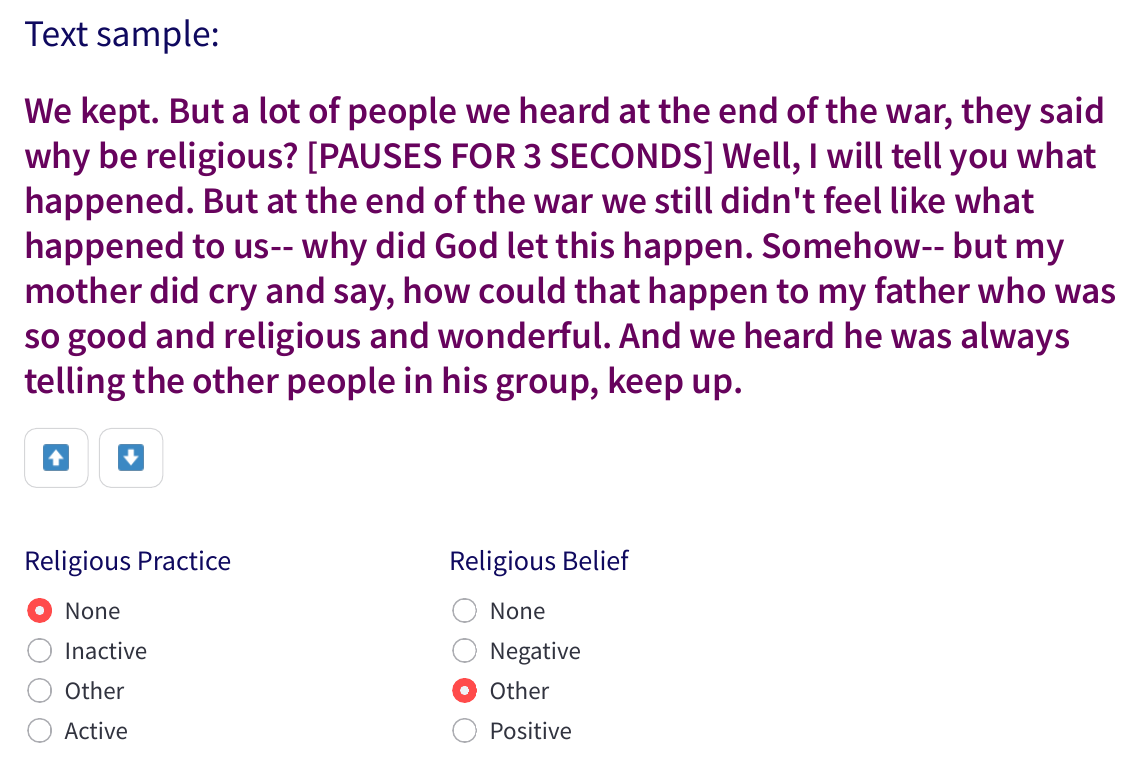}
  \caption{An annotation example from the platform we provided the annotators with to identify the survivor's valence of religious practice and belief in each segment.}
  \label{fig:platform}
\end{figure}

A segment describing a practice can be annotated with one of the classes -- \textit{Active}, \textit{Inactive} or \textit{Other}. The first two classes identify descriptions of participating in or abstaining from religious practices while \textit{Other} represents descriptions that do not meet the criteria of \textit{Active} or \textit{Inactive}, or match both of the classes simultaneously. For belief, the classification schema is similar, with the classes being \textit{Positive}, \textit{Negative}, and \textit{Other}. 
An example of the annotation process is illustrated in \autoref{fig:platform}, see \autoref{sec:appendix_annotations2} for the full guidelines. 

To construct the dataset for this task, we first run the religious content classifier to identify segments of interest. Since the majority of segments have no religious content, this step serves as initial filtering before the detailed annotation. 
Additionally, we use GPT-4o\footnote{\url{https://platform.openai.com/docs/models/gpt-4o} version gpt-4o-2024-05-13} to identify samples that belong to the minority classes, specifically the Belief classes and \textit{Other-Practice}.
We then incorporate into the dataset all samples that were predicted to be in these minority classes. This approach helps balance the dataset by increasing the number of examples in the underrepresented classes, ensuring both positive and negative instances are well-represented. The resulting dataset is the one provided to the annotators. 

We collect 1,165 annotations and hold an overlap of 540 samples. The average IAA score for the second task is 0.48, indicating moderate agreement and suggesting that this classification task is challenging even for human annotators.
We divide the data into three splits with equal proportions of each class. The test split is completely sampled from the overlap, to ensure the reliability of our results.  \autoref{detailed_breakdown} summarizes the label distribution for both annotation tasks.

\section{Modeling Trajectories}
\label{sec:modeling}

\subsection{Experimental setup}

Extracting the religious trajectory from a segmented testimony consists of fine-tuning and prompting LLMs. The main steps are: (1) filter religious content, (2) identify practices and beliefs, and (3) measure the valence of the protagonist's expressed attitude toward religion.

The annotation process yields 674 positive samples with religious content and the rest negative. We fine-tune a RoBERTa-large \cite{Liu2019RoBERTaAR} classifier on this task, using a balanced dataset (1,348 samples), divided into three splits: train, validation, and test, by a ratio of 0.8, 0.1 and 0.1, respectively.\footnote{Hyperparameters: 4 epochs, batch size=4, learning rate=1e-5, seed=5} To enhance the reliability of the model and reduce individuals’ bias in the evaluation, the entire test split was randomly sampled from the overlap data.

Evaluation of the model on the test set yields accuracy and recall scores of 97\% and 94\% respectively. Additionally, cross-validation over 10 random splits produces an average accuracy and recall of 91\%. 

Given the goal of this assignment to identify segments that form a node in the trajectory -- i.e., contain relevant practice and belief descriptions, these results demonstrate that the RoBERTa classifier is a reliable tool for this purpose.  

For modeling the trajectories, we prompt a series of GPT-4.1 models\footnote{\url{https://platform.openai.com/docs/models/gpt-4.1}, version: 2025-04-14} on multiple settings and compare their performance to fine-tuned LoRA adapters for Mistral-7B-instruct, using data from the second annotation task. \autoref{sec:mistral_gpt_prompts} contains the prompts we experiment with. Model selection is based on the macro average F1 score on the test set. For training, we use Mistralai's mistral-finetune repository\footnote{\url{https://github.com/mistralai/mistral-finetune}} with their default hyperparameters and test multiple seeds\footnote{Hyperparameters: Lora Rank=64, sequence length=16,384, batch size=1, learning rate=1e-4 number of micro-batches=1. Number of training steps=100, weight decay=0.1, pct-start=0.05. Seed for belief=1; for practice=5}. The best performing model across all practice and belief classes is GPT4.1 with an average F1 of 0.55 for practice and 0.64 for belief. The fine-tuned Mistral-7B scores 0.43 and 0.45 for practice and belief, respectively. In addition, we experiment with GPT-4.1-mini, which results in 0.48 (practice) and 0.62 (belief). See \autoref{model_selection} for the complete details on the multiple settings. 
Owing to limitations with large-scale API access, we use Mistral-7B to generate the full dataset of trajectories. The prompts we run follow the format of Self-Consistency \cite{Wang2022SelfConsistencyIC} and are carefully designed with guidance from Anthropic's prompt generator.\footnote{\url{https://console.anthropic.com/login}}
Considering the complicated annotation process and IAA, along with the models' bias toward stereotypes in the context of religion \cite{plaza-del-arco-etal-2024-divine}, these moderate F1 scores are unsurprising.                

While memorization is a valid concern when applying pre-trained LLMs, the specific combination of Holocaust testimony data with our labels is unlikely to have been seen in this exact form during pre-training and therefore is not expected to limit the generality of our results.

\subsection{Evaluation}

As we are not aware of any directly comparable reference for the trajectories. We therefore evaluate them against two reliable proxies, albeit not directly comparable ones: topic-modeling-based and thesaurus-based references.

\paragraph{Topic modeling based reference.}
\citet{ifergan} applied topic modeling using BERTopic \cite{grootendorst2022bertopic} to the same set of testimony transcripts. Their results include multiple topics that correspond to religious practices and beliefs.  Using this prior knowledge, we define a mapping of the topics to the practice/belief labels, and treat the result as reference trajectories. 
The segments that were used for topic modeling are about three times longer than the segments we use, leading to relatively sparse sequences. 
The induced topics that we compare against are provided in \autoref{topics_table}.

\paragraph{Thesaurus-based reference.} 
The testimony transcripts from the SF were divided into segments of one minute each,
indexed with a subset of keywords from the SF’s detailed thesaurus of ~8,000 terms.\footnote{\url{sfi.usc.edu/content/keyword-thesaurus}} 
We leverage the terms related to religious practices and beliefs, and map them to practice/belief labels to extract reference trajectories.
The full list of addressed terms can be found in \autoref{sec:list_of_terms}.

\begin{figure}[t]
    \centering
    \includegraphics[width=1\linewidth]{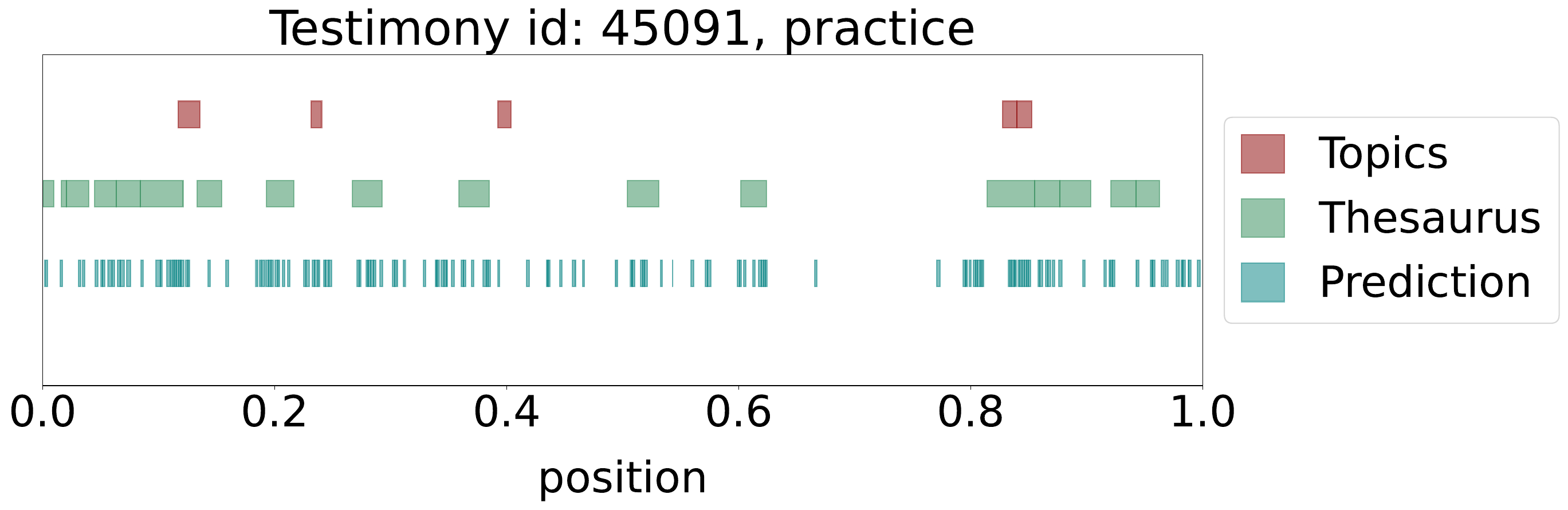}
    \caption{Alignment of predicted trajectories with reference trajectories for testimony ID: 45091, illustrating the differences and overlaps between them. The colored rectangle widths correspond to the segment lengths. The x-axis represents the normalized position within the testimony transcript.
    }
    \label{fig:ref_alignment}
\end{figure}

Both references have several limitations. First, the segments in the reference trajectories are larger than ours, leading them to contain information irrelevant to the practice and belief labels. Second, the reference sequences are sparse compared to the predicted ones. For the topics, our analysis of the topic model selects a single topic for each segment, rather than all the top relevant topics. For the thesaurus, despite the large number of terms, many segments lack labels for their matching terms. This may be a result of the exceedingly large set of keywords,
which may have resulted in recall issues in the SF annotation. In addition, the label set of the references is partial, meaning that the majority of the topics and terms we address do not have a specific valence. Considering these points, the false positive rate of the predictions is unknown, making the references suitable only for measuring recall. An example illustrating the alignment of the references and predictions is shown in \autoref{fig:ref_alignment}.

We run the fine-tuned Mistral-7B model on the full dataset, producing 1000 trajectories. In this setting, relying on our fine-tuned classifier, only religious content predictions are prompted, and not every single one of the original dataset segments. 
For each of the reference and predicted sequences, we quantify their distance.
For a given non-empty trajectory $T$ and a reference path $R$:

$$
min\_sum\_dist(T, R) = \sum_{r \in R}min_{t\in T}(|t-r|)
$$

If $T$ is empty, return the number of segments in $R$, and if $R$ is empty, return $0$.

We compare the sum of the distances to baseline trajectories. These baselines are artificial sequences designed by using our prior knowledge about the distribution (\autoref{fig:rc_dist}) of religious content throughout the dataset; each is the same length as the predicted sequence. The different baselines for a given class are defined in \autoref{baselines}.

\begin{table}
    \centering
    \small 
    \begin{tabularx}{\columnwidth}{p{0.4\columnwidth} X}
        \hline
        \textbf{Baseline} & \textbf{Definition} \\ \hline
        Equal & Scatter the points evenly between 0 and 1.  \\ \hline
        Original & Randomly select points from the original distribution of the predicted label.  \\ \hline
        Edges and middle & Randomly select points from three equal splits of (0, 1); each third contains the same portion of points that it has in the predicted distribution of the label. \\ \hline
        G-Edges and middle & Same as Edges\&middle except we sample the point from the normal distribution. \\ \hline
        2-Gaussian & All of the points sampled normally from the first half, same for the second half. \\ \hline
    \end{tabularx}
    \caption{The different baseline definitions for evaluating the predicted trajectories.}
    \label{baselines}
\end{table}

\section{Trajectory Clustering}

\begin{figure*}[ht]
    \centering
    \includegraphics[width=\linewidth]{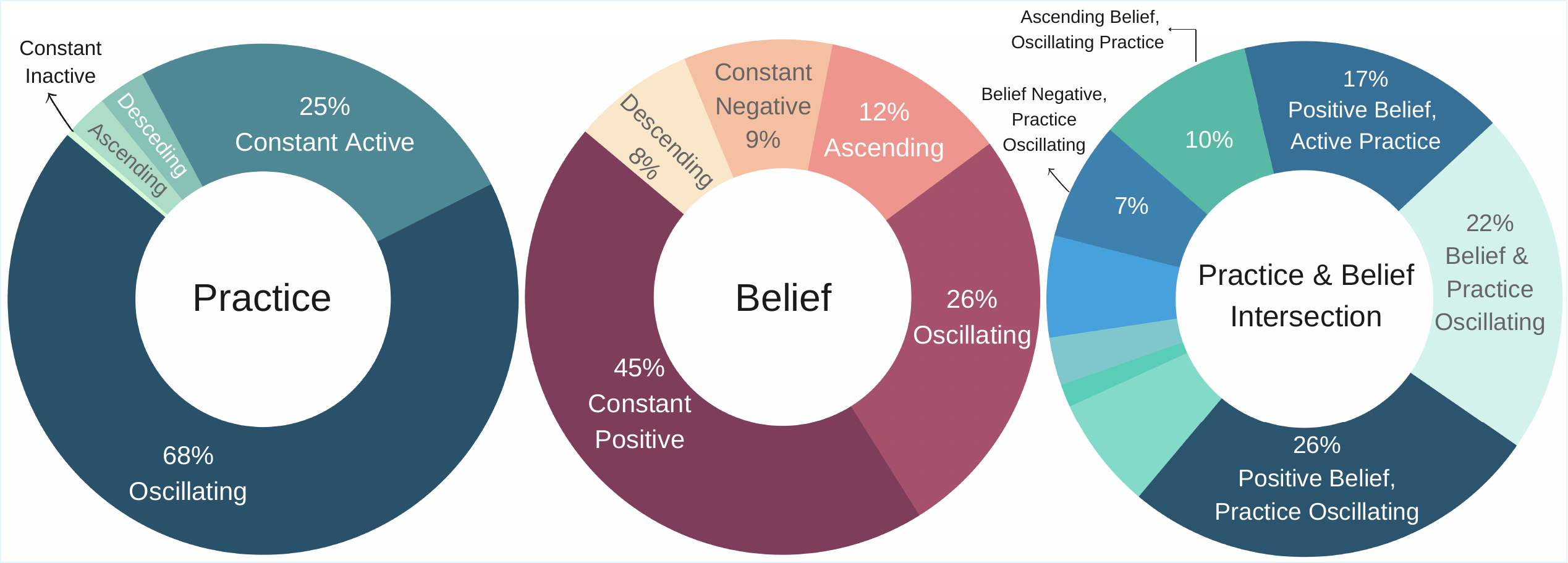}
    \caption{Religious Trajectory structure distributions, from left to right: 68\% of the practice trajectories have an oscillating structure and 25\% are constant-active.
    The belief trajectories have a similar distribution, with the majority of trajectories sharing a constant-positive structure (45\%) and the rest distributed among oscillating (26\%), ascending (12\%), constant-negative (9\%), and descending (8\%). For the intersection of the two aspects, the two large groups cover 44\% of the intersection, all have an oscillating practice structure, while the belief valence distributes evenly between oscillating and constant-positive structures.}
    \label{fig:structure_distributions}
\end{figure*}
    
After extracting the trajectories, we cluster them in two ways: 
according to a predefined taxonomy and using unsupervised methods. We expect to identify a few main clusters, each representing a different trajectory type that can be identified by generating a path that is more similar to each of the sequences within its cluster than to any of the other sequences. 


\subsection{Predefined Taxonomy}

We divide the trajectories into classes based on their valence and structure: {\it constant positive/negative, ascending/descending, oscillating}, and the degree of coverage, relative to the testimony storyline: $t_{max} - t_{min}$ (\autoref{structures}).
For this evaluation, each trajectory is filtered by removing any neutral values it contains and then shrunk by replacing a sequence of constant consecutive values with a single value. For example, [-1,1,0,1,1] was shrunk to [-1,1].

\begin{table}[ht]
    \centering
    \small 
    \begin{tabularx}{\columnwidth}{l X}
        \hline
        \textbf{Structure} & \textbf{Filtered\&shrunk series} \\ \hline
        Constant-Negative/Inactive & [-1]  \\
        Constant-Positive/Active & [1]  \\
        Ascending &  [-1, 1] \\
        Descending &  [1, -1]  \\
        Oscillating &  $ \{1, -1\}^n $ \\
        \hline
        \hline
        \multicolumn{2}{c}{\textbf{Coverage Level Definitions}} \\
        \hline
        Low & $ t_k - t_1 \leq 0.33 $ \\
        Medium & $ 0.33 \leq t_k - t_1 \leq 0.67 $ \\
        High & $ 0.67 < t_k - t_1  $ \\
        \hline
    \end{tabularx}
\caption{Definitions of the structures and coverage levels for the predefined taxonomy.}
\label{structures}
\end{table}

\subsection{Unsupervised Hierarchical Clustering}
\label{sec:h_clustering}

For hierarchical clustering, we use Dynamic Time Warping \cite{Berndt1994UsingDT} with multiple window sizes to measure the distances between the trajectories, and then run Agglomerative \cite{ZepedaMendoza2013} and HDBSCAN clustering \cite{McInnes2017hdbscanHD}.

\section{Results}

We evaluate the predicted trajectories against the references and present the results in \autoref{seq_eval}. Almost every reference point aligns with a predicted one, validating the predictions' recall. 

Besides the evaluation of the trajectories produced by promoting only religious content predictions, we randomly sample 101 testimonies, which we prompt and evaluate on the full set of segments. We observe that prompting on all segments, as opposed to just religious content, leads to a significant over-prediction rate for each class. See \autoref{sec:evaluation_subset} for the full analysis.

\begin{figure}[ht!]
    \centering
    \includegraphics[width=1\linewidth]{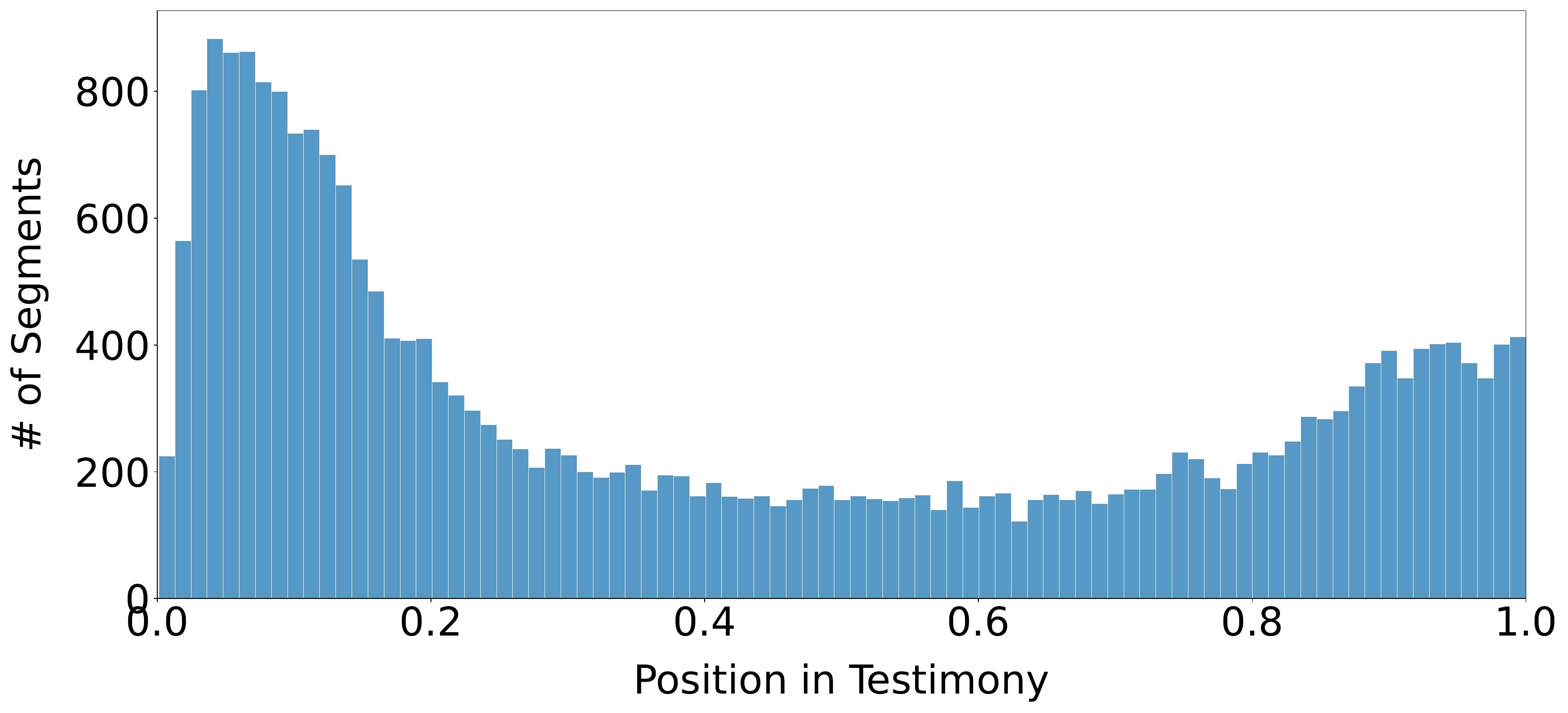}
    \caption{Distribution of all religious content, the distributions that we based the baselines on are according to each label separately.}
    \label{fig:rc_dist}
\end{figure}

\begin{table*}
    \centering
\begin{tabularx}{\textwidth}{l >{\centering\arraybackslash}X >{\centering\arraybackslash}X>{\centering\arraybackslash}X|| >{\centering\arraybackslash}X >{\centering\arraybackslash}X >{\centering\arraybackslash}X >{\centering\arraybackslash}X >{\centering\arraybackslash}X}    \hline
    & \multicolumn{3}{c||}{\textbf{Topic}} & \multicolumn{5}{c}{\textbf{Thesaurus}} \\
    & \multicolumn{1}{c}{\textbf{B}} & \multicolumn{1}{c}{\textbf{P}} & \multicolumn{1}{c||}{\textbf{P\textsuperscript{-}}} & \multicolumn{1}{c}{\textbf{B}} & \multicolumn{1}{c}{\textbf{P}} & \multicolumn{1}{c}{\textbf{P\textsuperscript{+}}} & \multicolumn{1}{c}{\textbf{B\textsuperscript{+}}} & \multicolumn{1}{c}{\textbf{B\textsuperscript{-}}} \\
    \hline
    \hline
    
Predicted           & \textbf{5} & \textbf{10} & \textbf{47} & \textbf{5} & \textbf{8} & \textbf{11} & \textbf{9} & \textbf{90} \\
Original scatter    & 33          & 40          & \textbf{47}          & 29             & 34         & 50          & 11          & 105          \\
Edges \& middle     & 19          & 29          & 59          & 19             & 28         & 29          & 16          & 99          \\
Equal scatter       & 21          & 20          & 60          & 19             & 17         & 21          &  \textbf{9}          & 142          \\
Normal-original     & 16          & 57          & 52          & 18             & 47         & 53          & 17          & 104          \\ 
\hline
\hline
\# Reference paths  & 335         & 905         & 187         & 282            & 787        & 761         & 98          & 217          \\  
\# predicted paths  & 954         & 998         & 742         & 954            & 998        & 990         & 796         & 480          \\
\# Reference points & 456         & 2,768       & 301         & 439            & 2,434      & 2,214       & 171         & 253          \\
\# predicted points & 6,051       & 23,274      & 1,987       & 6,051          & 23,274     & 20,086      & 3,204       & 917          \\ 
\hline
\hline
\end{tabularx}
\caption{Sum of $min\_sum\_dist$ for fine-tuned Mistral-7B predictions and baseline trajectories on the full dataset. \textbf{Original:} Sample from the distribution of the predictions. \textbf{Edges\&middle:} Random sample from three equal splits of (0, 1); according to the predicted distribution of the label.
\textbf{Equal:} Even scatter. 
\textbf{Normal-original:} Sample from the Gaussian with the variance and mean values of the predictions.
}
    \label{seq_eval}
\end{table*}

\subsection{Clustering}
Among the trajectories extracted by the pipeline, 83\% contain at least two points for both practice and belief. The taxonomy reveals that, for the belief aspect, 45\% share a constant positive shape, 26\% oscillate, and the rest are distributed among ascending (12\%), constant negative (9\%), and descending (8\%). 
For the positive trajectories, 58\% cover a substantial portion of the plot, whereas negative trajectories are sparser, 49\% covering up to two-thirds of the narrative. Practice trajectories have two large structures, 68\% oscillate and 25\% follow a constant-active pattern. Over 82\% of the two main structures have high coverage levels.
Examples for unsupervised clusters are given in \autoref{hierarchical_clustering}.

When comparing the trends according to the two aspects, the three most common combinations of structures are: Positive belief \& Practice Oscillating (26\%), both aspects Oscillating (22\%), and Positive belief \& Active Practice (17\%). 
These distributions are shown in \autoref{fig:structure_distributions}, further analysis in \autoref{bins}. 

\begin{figure*}[t]
    \centering
    \begin{minipage}[b]{0.4\linewidth}
        \includegraphics[width=\linewidth]{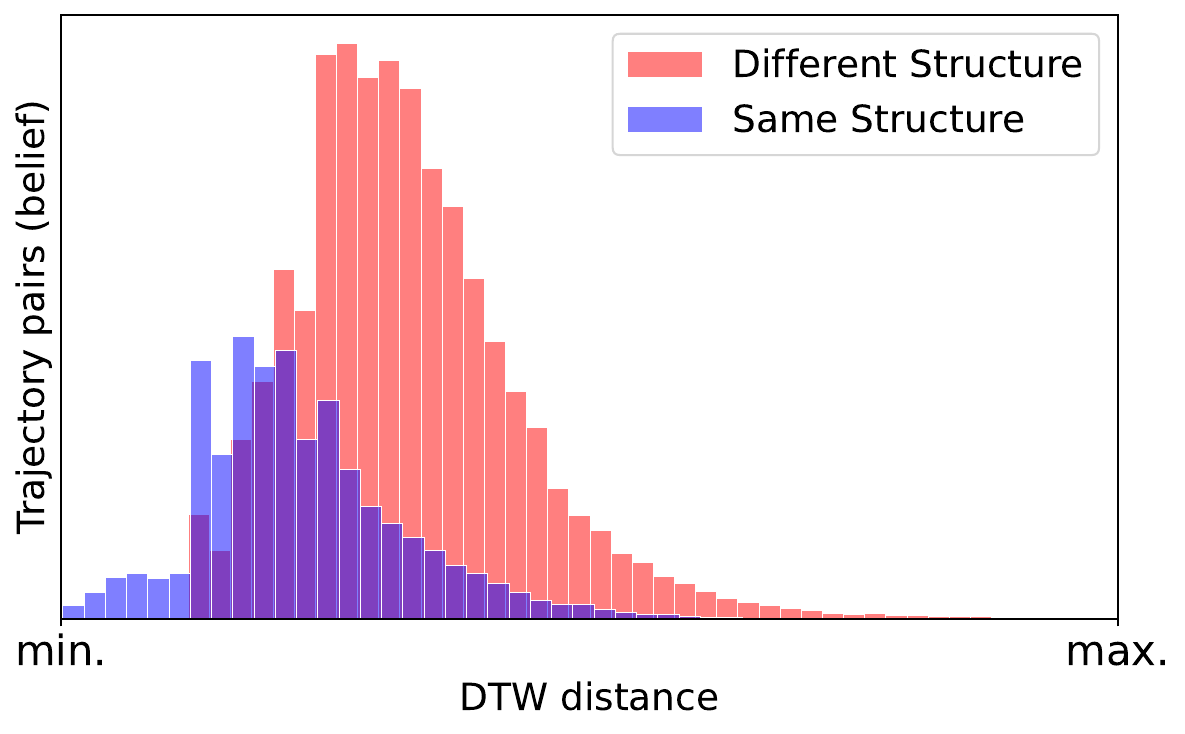}
        \label{fig:sub1}
    \end{minipage}
    \begin{minipage}[b]{0.4\linewidth}
        \includegraphics[width=\linewidth]{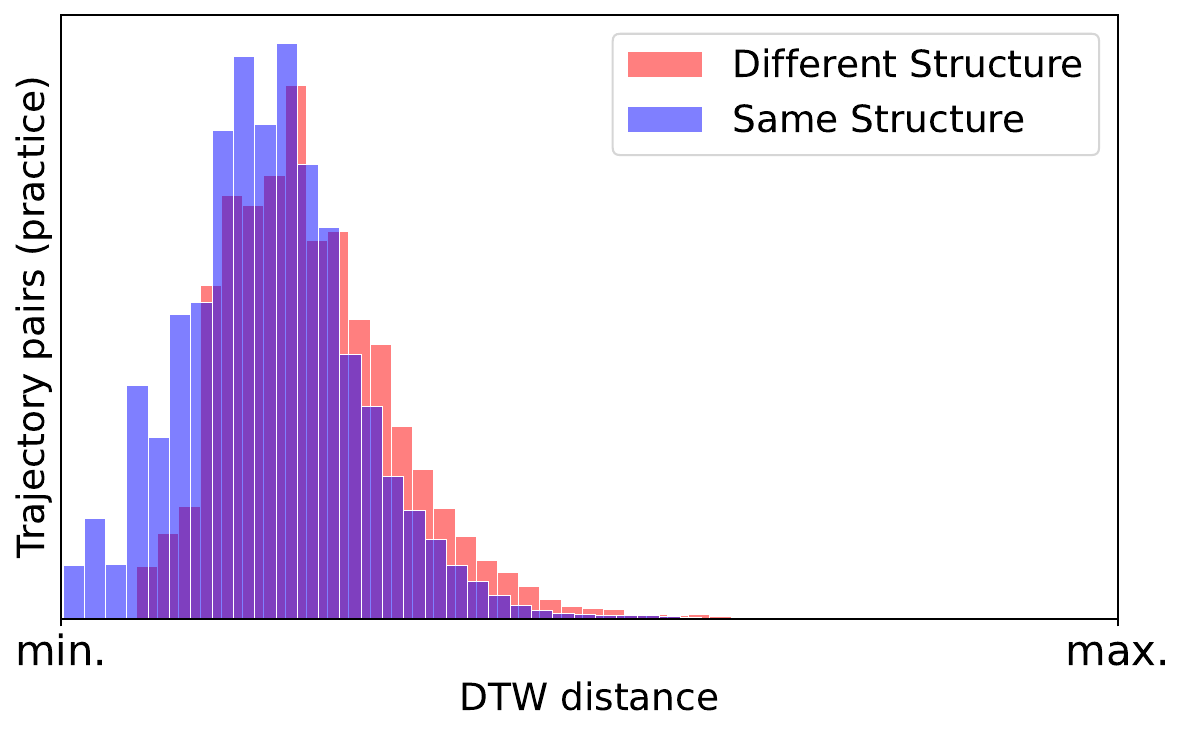}
        \label{fig:sub2}
    \end{minipage}
    \vspace{-0.6cm}
    \caption{DTW distance distributions of trajectory pairs: the left plot represents belief and the right practice. Blue indicates pairs with the same structures, and red indicates pairs with different structures. The distribution for identical structures is shifted toward smaller DTW distance values.
    }
    \label{fig:dtw_distributions}
\end{figure*}

\vspace{5mm}
\subsection{Clustering Evaluation}

We evaluate the clusters against human judgment, using a subset of trajectories with a length of at least 4 and a span greater than 0.75 of the testimony's length.

\paragraph{IAA.}
Annotators reviewed text segments from 121 trajectories and performed two tasks -- classifying them according to the taxonomy and identifying the more similar pairs from a list of triplets. IAA levels for classification were substantial for belief ($\alpha=0.66$) and moderate for practice ($\alpha=0.56$), while for the triplet task, IAA was moderate ($\alpha=0.58$ for belief, $\alpha=0.32$ for practice), showing that the task is challenging, even for humans.

\paragraph{Comparing human annotation with DTW.}
The DTW distances correlate with human judgment for similarity, validating its use for identifying similar trajectories. For similar pairs in the belief aspect, the average distance is 0.42 ($\sigma=0.19$), whereas for dissimilar pairs, it increases to 0.53 ($\sigma=0.18$), with a t-test revealing statistical significance ($p<0.006$). However, this pattern was not observed for the practice aspect. 

Taxonomy annotation shows that the DTW distance also correlates with the taxonomy's distinctions. The average distances are 0.38 (belief, $\sigma=0.14$) and 0.41 (practice, $\sigma=0.16$) for pairs of trajectories from the same class; while for pairs with different structures, the average distance is 0.43 (belief, $\sigma=0.14$), 0.44 (practice, $\sigma=0.16$). Running a t-test confirms statistical significance ($p<10^{-12}$ in both cases). This provides further validation for DTW by showing that part of the distance it captures is related to fundamental characteristics of the trajectory.

\paragraph{Manual and automatic taxonomy classification.}
Comparing the automatic and manual classification of trajectories into taxonomy classes, we observe F1 scores of 0.40 (belief) and 0.38 (practice) for both belief and practice.
Some discrepancies arise from the assumption that the predicted trajectories depend on the religious trajectory progressing along the narrative timeline; however, this does not hold for each individual case. 

As with manual taxonomy classification, automatic taxonomy prediction results correlate with DTW distances (\autoref{fig:dtw_distributions}). Pairs with the same structure exhibit smaller distances, with average distances of 0.11 (belief, $\sigma=0.05$) and 0.10 (practice, $\sigma=0.04$) for structures of the same class, compared to 0.16 (belief, $\sigma=0.06$) and 0.12 (practice, $\sigma=0.04$) for other pairs, reinforcing the link between structural alignment and DTW metrics.

\section{Discussion}

Our analysis demonstrates the value of this test case by identifying common structures of religious trajectories and grouping those with similar structures. For belief, the majority of trajectories present a constant positive structure, whereas for practice, most trajectories oscillate.

The distribution of religious content throughout the testimonies, demonstrating a common narrative sequence within the dataset, lends further support to the comparison of trajectories on the narrative timeline (\autoref{fig:rc_dist}). This can be caused by the SF interviewers' guidelines, which provide instructions for conducting the interviews.\footnote{\url{http://tiny.cc/sf-interview-guidelines}}

While these patterns reveal commonalities, interpreting them requires caution due to potential model biases. The confusion matrices (see \autoref{conf_matrices}) reveal that 64\% of belief statements with gold label Other were falsely predicted as Positive, and 56\% of practice–Other were misclassified as Active. This aligns with the idea of \textit{positivity bias} \citep{ashwin2023using, Fatahi2024ComparingEI}, where models tend to overpredict positive categories. In addition, because this work engages with the sensitive domain of religion, it is important to acknowledge potential stereotype biases rooted in pretraining data, which may introduce unintended sentiments \cite{elsafoury2023origins, garcia2023if}. For example, a Christian prayer (e.g., addressed to ``Dear Lord'' or ``Mother Maria'') is often predicted with a positive valence, even though our annotation guidelines define any non-Jewish religious expressions as carrying a negative valence.

Other directions we aim to cover include examining the testimonies that share structural similarities, exploring additional commonalities, and how these may correlate with metadata such as age, gender, pre/post-war hometowns, and experiences during the war. Any analysis will need to consider the general distribution of the data, as most survivors in our sample were born in Poland, with most interviews conducted in the United States and Canada.

Another perspective expected to contribute to a more nuanced picture of the patterns would include the non-positive-or-negative parts of the trajectory in the structure taxonomy analysis. The importance of incorporating these segments arises from the unsupervised clustering that induces a distinction between trajectories with multiple non-positive or negative segments, suggesting that these play a separate role in the trajectories. 

Additionally, inspecting the relationship between practice and belief trajectories, investigating whether one follows the other's pattern or is a shift of the other, and whether one can be predicted based on the other is an interesting avenue for future work.

\section{Conclusion}

This paper explores character development in narrative text, by analyzing religious trajectories from a dual perspective. We develop a framework to extract and cluster trajectories from 1,000 Holocaust testimony narratives. These findings encourage further research on utilizing LLMs to capture character development and thematic trajectories and offer valuable material for historical and sociological research. We believe that alignment with relevant literature can lend insights to the modeling.

\section*{Ethical Considerations}

According to the guidelines given by the SF archive, although the testimonies were not given anonymously, no identifying details are included in our analysis. Our codebase and scripts will be released on request, but they will not contain any data from the archives. The data and trained models used in our work will not be shared with third parties without the archives' consent. Permission can be requested from the SF archive to browse and research the testimonies.

\section*{Limitations}

As for the framework's limitations, it is important to consider the existence of reporting and survivorship bias in existing historical texts \cite{chen2024surveying}. Our data is limited to the interview transcripts, which capture the survivors' descriptions of their beliefs told in hindsight, without offering a broader understanding of their theological positions. Although the SF recorded testimonies in several languages and countries, it is important to note that our investigation only covered testimonies given in English.

We examine the trajectories without accounting for differences in their lengths and frequencies. However, these could provide a more nuanced understanding of the religious, social, and theological aspects reflected in the trajectories. Omitting the blank trajectories from our study might overlook a significant aspect. The absence of a religious trajectory may result from either a low level of religiosity or religion not being mentioned during the interview, which conveys a story, though the narrative behind it remains unknown. 

Additionally, much of this work relies on human annotators. Despite their shared professional backgrounds, each annotator brings different prior knowledge, which can affect agreement levels. For the cluster similarity evaluation, our analysis is restricted to the subset of testimonies that the annotators reviewed and analyzed.

\section*{Acknowledgments}
The authors acknowledge the USC Shoah Foundation -- The Institute for Visual History and Education for supporting this research. We thank Dr. Naama Seri-Levi, and Dr. Mali Eisenberg for their valuable insights, and our team of annotators for their research assistance. Grants from the Israeli Ministry of Science and Technology, the Israeli Council for Higher Education, and the Alfred Landecker Foundation supported this research.

\bibliography{custom}

\appendix

\section{The relationship between the chronological order of the testimonies and the narrative timeline}
\label{chronological_time}

\begin{figure*}[ht!]
    \centering
    \begin{minipage}[b]{0.4\linewidth}
        \includegraphics[width=\linewidth]{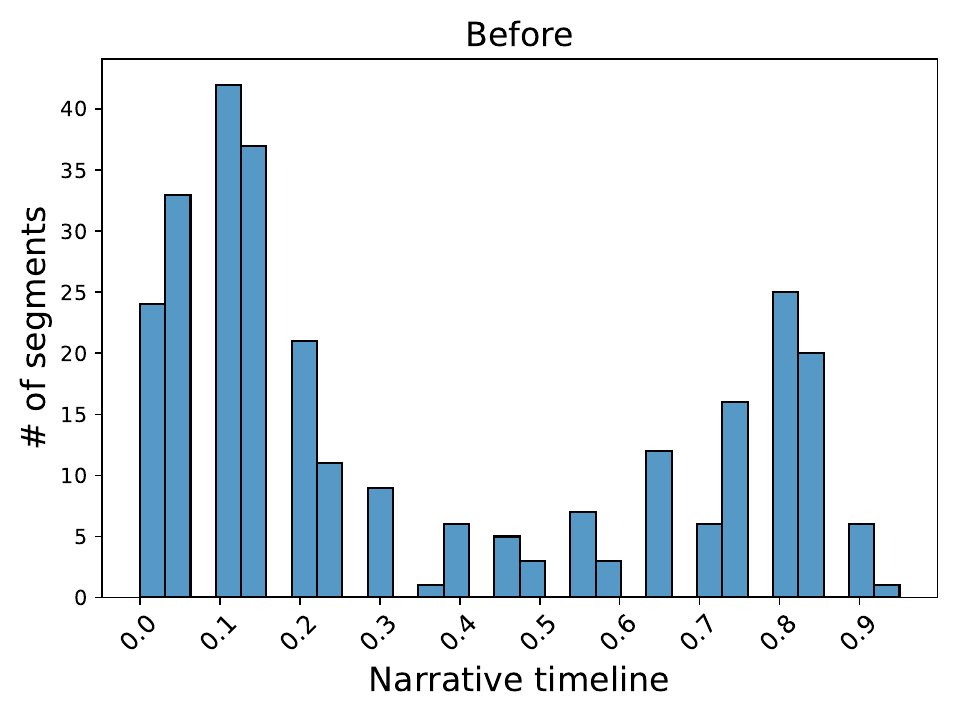}
        \label{fig:sub_1}
    \end{minipage}
    \begin{minipage}[b]{0.4\linewidth}
        \includegraphics[width=\linewidth]{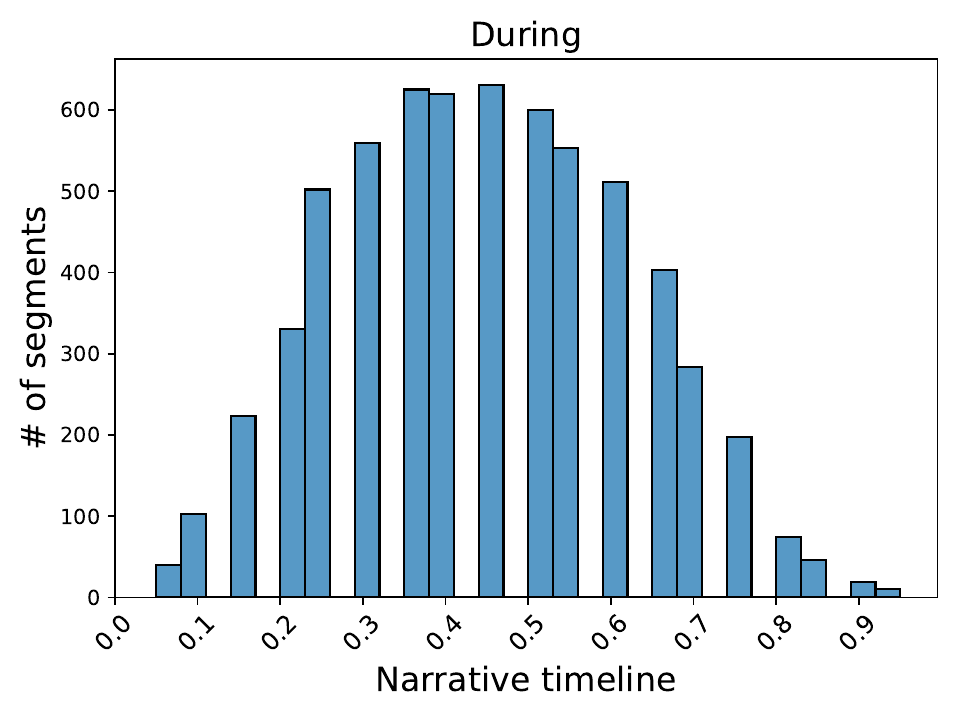}
        \label{fig:sub_2}
    \end{minipage}
    \begin{minipage}[b]{0.4\linewidth}
        \includegraphics[width=\linewidth]{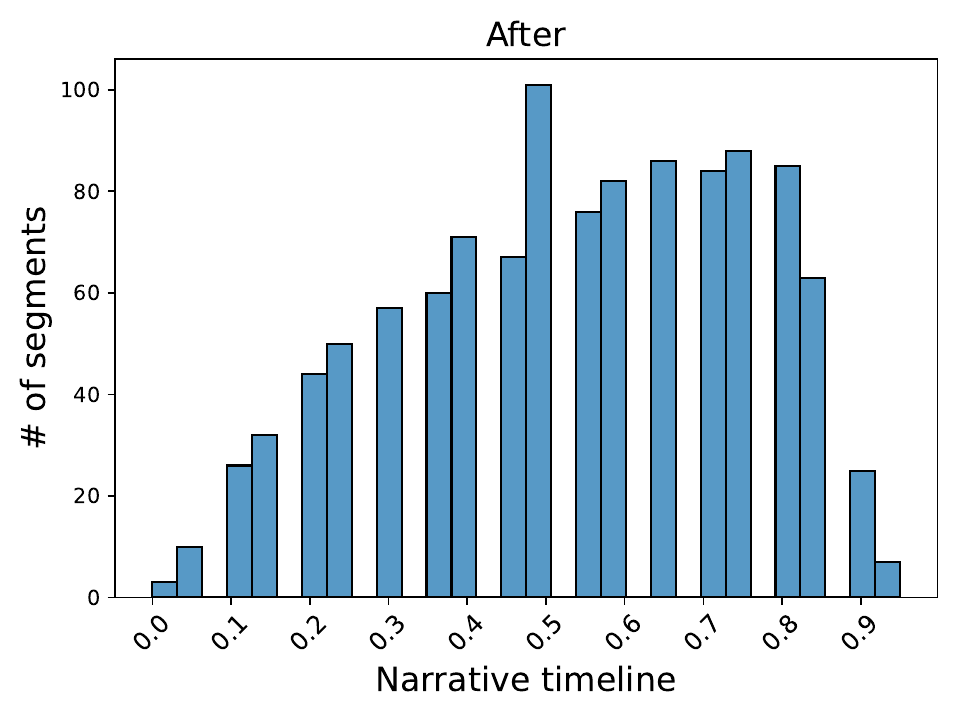}
        \label{fig:sub_3}
    \end{minipage}
    \begin{minipage}[b]{0.4\linewidth}
        \includegraphics[width=\linewidth]{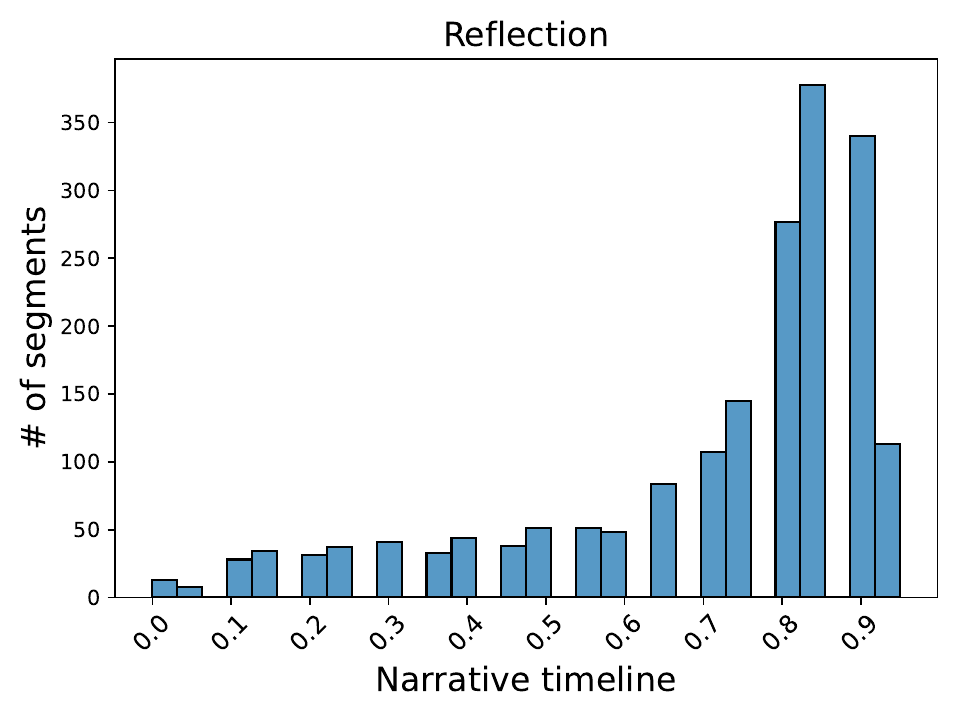}
        \label{fig:sub_4}
    \end{minipage}
    \caption{Distributions of the segments that match keywords mapped to periods of: before, during, and after the war, and reflection descriptions, along the narrative timeline}
    \label{fig:timeline_distributions}
\end{figure*}

\citet{wagner-etal-2022-topical} mapped the SF thesaurus keywords into periods of: before, during, and after the war, and reflection descriptions. For each segment in the data that matches one of these periods, we recorded its average position on the narrative timeline and got the pattern of: [before: 0.35, during: 0.45, after: 0.55, reflection: 0.71]. Both the distribution of the segments that match these keywords along the narrative timeline and the distribution of the religious content validate the comparison of the trajectories along the narrative timeline (see \autoref{fig:timeline_distributions}). 

The SF interviewers' guidelines include discussing pre-war religious activities at the beginning, and the survivor's present views toward the end, which strengthens this observation. In addition, the lengths of the responses are longer in the middle than at the beginning and the end, suggesting that the middle of the testimonies contains stories that the witness accounts in his choice of order. 

Regarding the structure of belief trajectories, particularly the meaning of oscillations, while belief descriptions can convey meaning about how belief is talked about, many of the belief segments reflect past beliefs, for example, \textit{“Every night I would take my prayer book… and cry”}. However, there is always tension between one's personal life as reflected in conversation in the present and their historically lived experience \cite{rosenthal2006narrated}. In this analysis, we concentrate on the former.

\section{Religious Content Annotation Guidelines}
\label{sec:appendix_annotations}
We are working on exploring changes in religious beliefs as reflected in testimony transcripts of Holocaust survivors; through looking at Jewish practices and beliefs described in the text. We want to capture all data segments from the testimonies that hold to the definition of religious content, we will label “TRUE” all segments that describe any Jewish religious practices or beliefs of the interviewee, \textbf{as well as segments that indicate the absence of religious practices or beliefs.} In detail:
\begin{itemize}
    \item 
    Religious practices and beliefs: Content that describes \textit{“Who we were, what we did, and what we believed in”}. For instance phrases in the style of: 
    \begin{itemize}
        \item 
        \textit{“We were orthodox” / “We were not religious”}
        \item 
        \textit{“My family belonged to a synagogue” / “We didn’t keep the Sabbath”}
        \item 
        \textit{“We hosted Pesach Seder” / “I wasn’t familiar with the Jewish holidays” }
    \end{itemize}
    \item 
    Descriptions of the interviewee's theological inner life, giving the reader information about their faith in God. For example:
    \begin{itemize}
        \item 
        \textit{ “I looked up to the sky, there must be a God above”}
        \item 
        \textit{“I didn't believe that God can look at this cruelty and all these dreadful things”}
    \end{itemize}
    \item 
    If the sample is written in first person (We, us, I), it is telling the story of the interviewee or their family and not stories about others, and its content meets our description of religious practices and beliefs then it will be labeled True.
    \item 
    We will consider samples written in the third person, only if they reflect on the interviewees' beliefs, or describe the environment that he grew up in.
    \begin{itemize}
        \item 
        \textit{“My parents explained that we were Jewish, secular Jews, nonbelievers”} - True
        \item 
        \textit{“My elder brother Rabbi Yosef Meyer was with me”} - False
        \item 
        \textit{“I think my brother probably was affected because he changed his religion, he became a Roman Catholic.”} - True, since implies that the speaker wasn’t affected.
    \end{itemize}
    \item 
    The phrases “Baruch Hashem '' and “Thank God '' are religious phrases and, therefore will be labeled True.
    \item 
    Zionism descriptions will be considered False; since the Zionist movement saw the land of Israel and the Hebrew language as integral parts of the Jewish national heritage, and not necessarily of religious significance.
    \item 
    Descriptions of Jewish Identity will be labeled False. We are focusing on \textbf{religious Identity and not Jewish Identity}, therefore we want to capture only segments that contain religious practices and beliefs, not signs of Jewish culture.
    \item 
    Jewish food or music, or speaking Yiddish - Not considered religious content
    \item 
    Hnnukah candles, Bris, Bat Mitzvah celebrations, and Friday night dinner - are considered positive religious content.
    \item 
    Getting a Jewish education
    \begin{itemize}
        \item 
        Secular Jewish education, such as attending Hebrew School, or a Tarbut institution, or participating in secular youth movements - Not considered religious content
        \item 
        Studying at Yeshivas or in any Jewish religious school, Bnei Akiva or Mizrahi - Positive religious content. 
    \end{itemize}
    \item 
    Mentions of miracles
    \begin{itemize}
        \item 
        If reflecting a belief will be labeled True, and not when used as a phrase.
        \item
        It is suggested to view the wider context of segments that mention miracles.
    \end{itemize}
    \item 
    Description of friendship relations with the non-Jewish population will be considered False, unless it indicates a religious approach.
    \begin{itemize}
        \item 
        \textit{“We were friendly with our gentile neighbors, though didn’t eat at their houses”} - True
    \end{itemize}
\end{itemize}

\paragraph{Some Examples}
\begin{enumerate}
    \item 
    \textit{“He gave me a Brucha, a blessing. I was very sick, a very sick child. And he told, don't worry. She'll grow. She'll be well, God will help her. ”}
    \item
    \textit{“he was able to survive, which it was a, a miracle. It was simply a miracle. Because, by rights, he was dying. He was on death's door”}
    \item
    \textit{“I got my presents under the Christmas tree, and then my girlfriend came to Hanukkah to my place. She got a present.”}
    \item
    \textit{    “I send them to the best yeshivas. This is the whole future of my life. And this was-- baruch HaShem-- always could be better. But I'm satisfied-- baruch HaShem. I accomplished.“}
    \item
    \textit{“He will always tell you, if you ask him, what's your religion, I am a Jew. And the boys really have neither been circumcised nor baptized. And it doesn't seem to have done them any harm so far.”}
    \begin{itemize}
        \item 
        Although it uses the word religion, The first part alone would not be considered Religious Content: “If you ask him, what's your religion, I am a Jew” since it is about Jewish identity, not Religion.
    \end{itemize}
    \item
    \textit{“I don't think God was absent, but I really don't understand yet.”}
    \item
    \textit{“A lot of kids thought that I am just a Catholic like they are. But it came a time, that when we came to school we had to stand up to say the prayers. And, of course, the Jewish children, we stood up, but we didn't say, and we didn't cross ourselves.” }
\end{enumerate}

\section{Practice and Belief Identification and Rating - Annotation Guidelines}
\label{sec:appendix_annotations2}
We are working on exploring changes in religious beliefs as reflected in testimony transcripts of Holocaust survivors; through looking at Jewish practices and beliefs described in the text. This task has two parts: Identifying the class of each sample and then determining the speaker's approach toward the specific class. 
\paragraph{Part 1 - Practice and Belief Identification}
Practice and Belief Identification is a multi-label classification task. The dataset for this task was randomly sampled from data classified as religious content by a machine learning classifier with high accuracy. The task is to map each sample to one or more of the classes or mark it as falsely recognized as religious content.

\textbf{Label Definitions}
\begin{itemize}
    \item 
    Practice - A ritual, action, or activity motivated by the Jewish religion. Also descriptions reflecting an absence of Jewish religious practices will be mapped to this label.
    \begin{itemize}
        \item 
        Not including descriptions of Jewish communities or culture. 
        \item 
        Examples: Saying prayers, going to synagogue, keeping Kosher.
    \end{itemize}
    \item 
    Belief - Inner life descriptions of ideas or thoughts related to God or religion. Including philosophy and feelings about God, and descriptions of miracles.
    \begin{itemize}
        \item 
        Excluding descriptions of Jewish identity or Zionism.
    \end{itemize}
    \item 
    Other - Samples that were falsely recognized as religious content 
\end{itemize}

A general note: We will consider samples written in the third person, only if they reflect on the interviewees' beliefs, or describe the environment that he grew up in.

\paragraph{Some Examples}
\textbf{Practice}
\begin{enumerate}
    \item 
    \textit{“He gave me a brucha, a blessing. I was very sick. He gave me a blessing. I was usually a very sick child-- wild, spoiled, and sick. And he told, don't worry. She'll grow. She'll be well, -- God will help her. ” 
    \begin{itemize}
        \item 
        Note: Although phrased in the third person this example is relevant since the narrator of the text went to get the Bracha. It is also considered Belief, since it contains the phrase “God will help her”.
    \end{itemize}
    \item 
    \textit{“ I got my presents under the Christmas tree, and then my girlfriend came to Hanukkah to-- to my place. She got a present.”}
}
\end{enumerate}

\textbf{Belief}
\begin{enumerate}
    \item 
    \textit{“He gave me a brucha, a blessing. I was very sick. He gave me a blessing. I was usually a very sick child-- wild, spoiled, and sick. And he told, don't worry. She'll grow. She'll be well, -- God will help her. ”
}
    \item 
    \textit{“he was able to survive, which it was a, a miracle. It was simply a miracle. Because, by rights, he was dying. He was on death's door”}
    \begin{itemize}
        \item 
        Note: While the text is written in the third person point of view, it reflects the narrators faith.   
    \end{itemize}
    \item 
    \textit{“I had the will, strong willpower to live. Every night I would take my prayer book. Every night somebody was in charge from the twins to look out. Each bunk beds, everybody is asleep, it's quiet. Then came my turn. One night, I would take my prayer book and cry. Say Psalms. It's so good. I felt close and came–”}
    \begin{itemize}
        \item 
        Note: this example is also considered Practice, since it describes a daily ritual.
    \end{itemize}
\end{enumerate}

\textbf{Other - Neither a practice or belief descriptions}
\begin{enumerate}
    \item 
    \textit{“Was there any particular reason for going to a Jewish school for secondary school? The reason was that I didn't want to sit in exam to go to grammar school. I was too frightened. It was a grave mistake.”}
\end{enumerate}

\paragraph{Part 2 - Rating Jewish Practice and Belief Text Samples}
After identifying the classes of a text sample, our task is to rate the narrator’s approach to the specific class on a scale from -1 to 1.

\textbf{Rating Definitions:}
\begin{itemize}
    \item 
    Active - Actively practicing a Jewish religious ritual
    \item 
    Inactive - Violating Jewish religious practices, descriptions of not observing, actively not practicing a Jewish religious ritual, or practicing a different religion.
    \item 
    Neither active or inactive valence of Jewish religious practice, or both simultaneously.   
\end{itemize}

\paragraph{Practice examples}
\textbf{Active}
\begin{enumerate}
    \item 
    \textit{“How would you describe your family's religious life? Orthodox.”}
    \item 
    \textit{“When did you have a Bar mitzvah? When I was 13. In the-- in the main synagogue.“}
\end{enumerate}

\textbf{Inactive}
\begin{enumerate}
    \item 
    \textit{“And did you-- did you observe Shabbat in any way, light candles, go to synagogue? No, no, no candles, no Shabbat.”}
    \item 
    \textit{“The church became a very important part of my life. Although we didn't go often, any time that I went with my Catholic family, the Leszkowicz or Maciejewski family, I felt safe. And this cross that you're looking at was given to me by my Catholic grandmother.”}
\end{enumerate}

\textbf{Other}
\begin{enumerate}
    \item 
    \textit{“Was there davening on the train? Yeah. Sure, they were davening.”}
    \item 
    \textit{“We were sleeping in an attic, in a cellar, here. If we were caught, that would be the end. Meanwhile, Rosh Hashanah-- and-- and don't forget.”}
    \begin{itemize}
        \item 
        We don’t know whether the speaker celebrated Rosh Hashana or is just mentioning the time od the year, therefor it’s valence is “other”. 
    \end{itemize}
    \item 
    \textit{“You said your family was not religious. So in what way did you identify with Judaism? I would say-- ethnically and community feeling. my grandmother and also my maternal grandparents, they went also only I think mainly Rosh Hashanah and Yom Kippur, not more than that.”}
    \begin{itemize}
        \item 
        On the one hand, going to synagogue on the high holidays - Active. \\
        On the other, not a religious family - Inactive. \\
        → Altogether the signal is “other”.
    \end{itemize}
\end{enumerate}

\paragraph{Belief examples}
\textbf{Positive}
\begin{enumerate}
    \item 
    \textit{“And my mother kept on saying, this is going to be the Passover of Passovers, because we were liberated. Moses took us out of Egypt. And now, the world is going to take us out of Egypt from Hitler. But it didn't happen. The Jews fought for a month there. They fought most-- the strongest army in the world. They, of course, at the end, lost.”}
    \begin{itemize}
        \item 
        Note: This is a practice example as well.
    \end{itemize}
    \item 
    \textit{“What did you do when you saw that being done to the children? You think I could go up and say don't do it? Nothing. I just prayed. I just prayed. But I had to look. When I look at this now, that's what I saw. That's what had happened. And still Germany is so strong. And Germany is a country.”}
        \item 
    \textit{“But I still belong-- believe in God. And if he's good, or if he can close his eyes, and then see us, and not to see and not to hear, but it comes a time when he takes his hands away, and from his eyes and ears, and he sees. And I hope that I will see, too. If I don't see, as my children, grandchildren will see and hear that he took his hands away from his eyes and ears, and he can help them.”}
\end{enumerate}

\textbf{Negative}
\begin{enumerate}
    \item 
    \textit{“Why not? How can I believe? How is it possible? I just keep the tradition for my children and grandchildren sake and my friends who don't even keep the tradition. So they come. And we have nice dinner. And we try to be in good mood, but nothing Jewish in us anymore, no.”}
    \item 
    \textit{“How were you mentally at this stage, Leah, after all your experience? Very angry, very bitter. I didn't believe there was a god. I didn't want to know such a thing as a god. Very guilty.”}
    \item 
    \textit{“And that was our religion. But we didn't keep Shabbat. We didn't keep kosher. Nothing. And during the time in Auschwitz, I didn't even believe in religion. I didn't believe that the God can look at this cruelty and at all these dreadful things. I didn't believe in God. Maybe it was just a shock to the-- I don't know.”}
    \begin{itemize}
        \item 
        Note: This is a practice example as well.
    \end{itemize}
\end{enumerate}

\textbf{Other}
\begin{enumerate}
    \item 
    \textit{“What do you think kept you going? What kept me going? Good question that, obstinacy-- I didn't want to die. And however much I didn't want to die, the decision didn't rest with me. This-- this is the interesting-- it didn't rest with me at all. Maybe God helped-- helped me to-- to somehow that Dr. Mengele didn't pick me every time he went through these selections. I don't know. I don't know. It certainly wasn't me.”}
    \item 
    \textit{“And what about your beliefs? Do-- do you practice Judaism? Do you believe in a deity? And is that in any way impacted by your experiences one way or the other? Well, I believe very much in the importance of what I would call tradition. And I think that our tradition is-- is something to be very proud of. I think we have a very rich tradition, we have very rich history, a very unique history in the world. And I think it is-- it is very much part of us.”}
    \begin{itemize}
        \item 
        The speaker doesn’t deny or assure faith in God.
    \end{itemize}
        \item 
    \textit{“What has-- have you inherited that's helped you survive it? Well, I have more questions, I mean, about God than my father does, I'll tell you that. But I do respect very much my father's faith in God and feel that, if he can live though that and believe in God, then I certainly don't have a right to question that. So I've learned that people can go through hell, still believe in God, and go on in, in most ways in their lives, go on and enjoy life, and enjoy children. My father loves young children. He loves children. His whole life is children.”}
\end{enumerate}

\label{sec:prompt_enlarge}

\section{Annotations}
\label{detailed_breakdown}
\autoref{breakdown_table}: The distribution of all the annotations collected.

\autoref{overlap_dist}: The label distribution for the overlap subset of annotations.

\begin{table*}[!ht]
    \centering
    \begin{tabularx}{\textwidth}{l l >{\centering\arraybackslash}X}
        \hline
        \textbf{Annotation Task} & \textbf{Class} & \textbf{Annotations Collected} \\
        \hline
        \hline
        \multirow{2}{*}{Religious content identification}
            & Positive & \textbf{674} \\
            & Negative & \textbf{4321} \\
        \hline
        \multirow{4}{*}{Practice valence classification}
            & Active & \textbf{343} \\
            & Inactive & \textbf{77} \\
            & Other-practice & \textbf{49} \\
            & None & \textbf{607} \\
        \hline
        \multirow{4}{*}{Belief valence classification}
            & Positive & \textbf{122} \\
            & Negative & \textbf{68} \\
            & Other-belief & \textbf{44} \\
            & None & \textbf{862} \\
        \hline
    \end{tabularx}
    \caption{Breakdown of annotations collected across three annotation tasks: (1) Religious content identification, (2) Practice valence classification, and (3) Belief valence classification.}
    \label{breakdown_table}
\end{table*}

\begin{table*}[ht!]
    \centering
    \begin{tabularx}{\textwidth}{l X c c}
        \toprule
        \textbf{Annotation Task} & \textbf{Label Distribution \newline(Agreed Upon Samples)} & \textbf{Agreement Size} & \textbf{Overlap Size} \\
        \midrule
        Religious content classification &
            Positive: 227 \newline Negative: 520 &
            747 &
            833 (21\%) \\
        \midrule
        Practice valence &
            Active: 84 \newline Inactive: 40 \newline Other: 32 \newline None: 189 &
            345 & 540 (46\%) \\

        \midrule
        Belief valence &
            Positive: 37 \newline Negative: 34 \newline Other: 34 \newline None: 314 &
            417 &  540 (46\%) \\
        \bottomrule
    \end{tabularx}
    \caption{Overlap label distribution for both annotation tasks. Agreement size is the number of pairs agreed open; Overlap size is the total number of overlapping samples between annotators.}
    \label{overlap_dist}
\end{table*}

\section{GPT-4o Prompt for enlarging the portion of the minority classes}
\paragraph{Belief Recognition}
\begin{quote}
Your task is to determine if the text contains a phrase describing a religious belief. Any length of selection is considered a phrase!

Class definitions:
BELIEF = One or more of the following: Spirituality, inner-life descriptions, beliefs about God, relationship with God, principles, identity, philosophy about God, feelings about God, Grace to God, thoughts about God, emotional experiences that come from beliefs about God, mentions of miracles. 
God == The Lord == Hashem.

Other = descriptions of anything else.

The answer must contain a single class, Do not add any words! Do not provide any explanations!
The only labels allowed are:
BELIEF
OTHER

Examples:

Text: “My mother was a very religious person and a righteous person, and she felt that perhaps it would be best that she go to live with her aunt. Her aunt kind of appealed to her religious fair being, I guess, if you will.”
Label: OTHER

Text: “And God should be with you. So my father said, OK, you're going to see. They're going to kill us from the back. From here, we're not going out alive. I said, look it, whatever God wants, this is what's going to be.”
Label: BELIEF

Text: “With my education that I had on the Zionist movement, I realized that they taught me what it means to be a Jew. And my Hebrew school taught me. And the Bible taught me. But then I realized what it meant to be a Jew.”
Label: BELIEF

Text: “And my mother kept on saying, this is going to be the Passover of Passovers, because we were liberated. Moses took us out of Egypt. And now, the world is going to take us out of Egypt from Hitler. But it didn't happen. The Jews fought for a month there. They, of course, at the end, lost.”
Label: BELIEF

Text: “And that's the way my mother went to Palestine as a tourist, my brother with the Youth Aliyah. We were big Zionists. ”
Label: OTHER

Text: “And you're proud today to be a Jew? Oh, my goodness. And my friend upstairs, I made a deal with him a long time ago. And he kept his side of the bargain, I must keep my side of the bargain, because he's given me miracles, and he's shown me he is there.”
Label: BELIEF

Text: “I want to believe that there is a purpose for all this. Sometimes things are very hard to accept, but I still believe.”
Label: BELIEF

Text: “It was the time between Pesach and, and, and Lag B'Omer. And we were supposed to get together for Lag B'Omer.”
Label: OTHER

Text: “My parents explained that we were Jewish, secular Jews, nonbelievers”
Label: BELIEF

Text: “My father's family was a fairly large traditional Jewish family. Orthodox. I remember my mother wearing a wig.”
Label: OTHER

Text: “I have more questions about God than my father does. But I do respect very much my father's faith in God and feel that, if he can live through that and believe in God, then I certainly don't have a right to question that. So I've learned that people can go through hell, still believe in God.”
Label: BELIEF

Text: “This is a prayer book that was my grandmother's. She carried it with her all through Europe, into Israel, and then she brought it to this country.”
Label: OTHER

Text: “I took her death like a punishment from God, Why am I being punished? I pray every morning. I was so hurt. I felt that God did it, because they said God does it. I was very angry at God.”
Label: BELIEF

Text: “They observe the Jewish holiday, but they weren't real Orthodox. They were very Conservative. To the shul they used to go only on the holidays.”
Label: OTHER

Text: “Simchas Torah they went over the roof dancing around. And like all the Hasidim doing.”
Label: OTHER

Text: “After all what my father went through, we could still believe in HaShem, God, and that we have to believe in Him and the continuity of the Jewish people.”
Label: BELIEF

Text: “How can I believe? How is it possible? I just keep the tradition for my children and grandchildren sake and my friends who don't even keep the tradition. but nothing Jewish in us anymore, no.”
Label: BELIEF

Text: “We went to church every day.”
Label: OTHER

Text: “It was like a miracle happened”
Label: BELIEF

Text: “Continue on the Jewish life. I'm very proud when I see my grandchildren say the Kiddush. I know my parents they must kvell, in heaven, to see this.”
Label: BELIEF

Text: “I had a two-tier program, one secular education, the other one is Jewish or Hebrew studies, which first included learning how to read, then Mishnah, Chumash, and Talmud.”
Label: OTHER 
\end{quote}

\paragraph{Practice Recognition}
\begin{quote}
Your task is to determine if the text contains a phrase describing a religious practice. Any length of selection is considered a phrase!

Class definitions:
PRACTICE = An action or activity connected to religion, Jewish community, or Jewish culture. Examples: Being orthodox, saying religious texts, going places connected to the Jewish religion, keeping certain restrictions, speaking Yiddish.

OTHER = descriptions of anything else.

The answer must contain a single class, Do not add any words! Do not provide any explanations!
The only labels allowed are:
PRACTICE
OTHER

Examples:

Text: “My mother was a very religious person and a righteous person, and she felt that perhaps it would be best that she go to live with her aunt. Her aunt kind of appealed to her religious fair being, I guess, if you will.”
Label: PRACTICE

Text: “I had a few friends that came back. We resumed life, normally, planning to eventually to go to Palestine or to Israel. The community grew to about 300 people after the war. ”
Label: OTHER

Text: “And God should be with you. So my father said, OK, you're going to see. They're going to kill us from the back. From here, we're not going out alive. I said, look it, whatever God wants, this is what's going to be.”
Label: OTHER

Text: “With my education that I had on the Zionist movement, I realized that they taught me what it means to be a Jew. And my Hebrew school taught me. And the Bible taught me. But then I realized what it meant to be a Jew.”
Label: PRACTICE

Text: “And my mother kept on saying, this is going to be the Passover of Passovers, because we were liberated. Moses took us out of Egypt. And now, the world is going to take us out of Egypt from Hitler. But it didn't happen. The Jews fought for a month there. They, of course, at the end, lost.”
Label: PRACTICE

Text: “And that's the way my mother went to Palestine as a tourist, my brother with the Youth Aliyah. We were big Zionists. ”
Label: PRACTICE

Text: “And you're proud today to be a Jew? Oh, my goodness. And my friend upstairs, I made a deal with him a long time ago. And he kept his side of the bargain, I must keep my side of the bargain, because he's given me miracles, and he's shown me he is there.”
Label: OTHER

Text: “I want to believe that there is a purpose for all this. Sometimes things are very hard to accept, but I still believe.”
Label: OTHER

Text: “It was the time between Pesach and, and, and Lag B'Omer. And we were supposed to get together for Lag B'Omer.”
Label: PRACTICE

Text: “My parents explained that we were Jewish, secular Jews, nonbelievers”
Label: PRACTICE

Text: “My father's family was a fairly large traditional Jewish family. Orthodox. I remember my mother wearing a wig.”
Label: PRACTICE

Text: “I have more questions about God than my father does. But I do respect very much my father's faith in God and feel that, if he can live through that and believe in God, then I certainly don't have a right to question that. So I've learned that people can go through hell, still believe in God.”
Label: OTHER

Text: “This is a prayer book that was my grandmother's.”
Label: PRACTICE

Text: “I took her death like a punishment from God, Why am I being punished? I pray every morning. I was so hurt. I felt that God did it, because they said God does it. I was very angry at God.”
Label: PRACTICE

Text: “They observe the Jewish holiday, but they weren't real Orthodox. They were very Conservative. To the shul they used to go only on the holidays.”
Label: PRACTICE

Text: “Simchas Torah they went over the roof dancing around. And like all the Hasidim doing.”
Label: PRACTICE

Text: “After all what my father went through, we could still believe in HaShem, God, and that we have to believe in Him and the continuity of the Jewish people.”
Label: OTHER

Text: “How can I believe? How is it possible? I just keep the tradition for my children and grandchildren sake and my friends who don't even keep the tradition. but nothing Jewish in us anymore, no.”
Label: PRACTICE

Text: “We went to church every day.”
Label: PRACTICE

Text: “It was like a miracle happened”
Label: OTHER

Text: “Continue on the Jewish life. I'm very proud when I see my grandchildren say the Kiddush. I know my parents they must kvell, in heaven, to see this.”
Label: PRACTICE

Text: “I had a two-tier program, one secular education, the other one is Jewish or Hebrew studies, which first included learning how to read, then Mishnah, Chumash, and Talmud.”
Label: PRACTICE
\end{quote}

\section{Instruction Prompts}
\label{sec:mistral_gpt_prompts}

The Instruction prompts we ran for GPT4o and Fine-tuning Mistral:\\
\paragraph{Prompt for Belief Classification} 

\begin{quote}
Your task is to carefully read this text and determine the speaker's valence of Jewish religious belief in God, based on the following classification system:
    POSITIVE: The text expresses the narrator's belief in God according to the Jewish religion, or his existing relationship with God.

    NEGATIVE: The text expresses the narrator's lack of belief in God according to the Jewish religion or a rejection of religious beliefs.

    AMBIGUOUS: The text expresses a relationship with God that does not meet the criteria of the classes POSITIVE or NEGATIVE. This includes questioning God while believing in his existence.

    NONE: The text does not directly imply the speaker's belief in God and religion or their lack of it. This includes texts written in the third person that do not describe the speaker's personal beliefs or family environment.

First, write out your reasoning for classifying the text inside <reasoning> tags. Consider the content and tone of the text, and how it aligns with the definitions provided above.\\
After writing your reasoning, output your final classification as a single word (POSITIVE, NEGATIVE, AMBIGUOUS, or NONE) inside <classification> tags.
\\
Use HTML tags in your response.

Do not add any words after </classification>.
\end{quote} 
    
\paragraph{Prompt for Practice Classification} 

\begin{quote}
    Your task is to carefully read this text and determine the speaker's valence of Jewish religious practice described in the text, if any, based on the following classification system:
    \begin{enumerate}
        \item 
        ACTIVE = The text expresses the narrator actively practicing a Jewish religious ritual.
        \item 
        INACTIVE = The text expresses the narrator violating Jewish religious practices or not observing/actively not practicing a Jewish religious ritual.
        \item 
        AMBIGUOUS = The narrator of the text expresses a Jewish religious practice, that does not meet the criteria of the classes ACTIVE or INACTIVE, or the text matches both of the classes at the same time.
        \item
        NONE = The text does not directly discuss the speaker participating in a religious practice or violating one. This includes texts written in the third person that do not describe the speaker's personal valence of practicing religion or family environment.       
    \end{enumerate}  
    
    First, write out your reasoning for classifying the text inside <reasoning> tags. Consider the content and tone of the text, and how it aligns with the definitions provided above. \\
    After writing your reasoning, output your final classification as a single word (ACTIVE, INACTIVE, AMBIGUOUS, or NONE) inside <classification> tags. \\
    Use HTML tags in your response. \\
    Do not add any words after </classification>.
\end{quote}

\section{Model Selection F1 scores}
\label{model_selection}
\autoref{model_scores}: A summary of the performance across models on the test data for rating practice and belief valence.

\begin{table}[ht!]
\centering
\small
\begin{tabularx}{\columnwidth}{l l X X}
\textbf{Model} & \textbf{Prompt ID} & \textbf{Practice} & \textbf{Belief} \\ \hline
GPT-4o            & few-shot   & 0.51          & 0.60          \\
Fine-tuned Mistral& zero-shot  & \textbf{0.43} & \textbf{0.45} \\
GPT-4o            & zero-shot  & 0.47          & 0.55          \\
Mistral           & zero-shot  & 0.26          & 0.25          \\
GPT-4.1           & few-shot   & \textbf{0.55} & \textbf{0.64} \\
GPT-4.1-mini      & few-shot   & 0.48          & 0.58          \\
GPT-4.1-nano      & few-shot   & 0.27          & 0.33          \\
GPT-4.1           & zero-shot  & 0.48          & 0.62          \\
GPT-4.1-mini      & zero-shot  & \textbf{0.48} & \textbf{0.55} \\
GPT-4.1           & 1-shot     & 0.50          & 0.62          \\
\end{tabularx}
\caption{Macro F1 scores for the various models and settings tested for selecting models to produce trajectories with}
\label{model_scores}
\end{table}

\section{Topic Model Based Evaluation}
\label{topics_table}
The topics we address, and their matching categories are in \autoref{topics}.

\begin{table}[ht]
    \centering
    \begin{tabularx}{\columnwidth}{X ll}
        \hline
        \textbf{Topic} & \textbf{Class} & \textbf{Sub-Class} \\ \hline
        Synagogue, holiday(s), Shabbos, religious, Shabbat, Shul, Friday, Passover, Jewish  & practice & - \\ 
        \hline
        Bar, Mitzvah, Torah, Synagogue, Mitzvahed, Shul, Rabbi, religious & practice & - \\
        \hline
        God, believe, religion/religious, faith, question & belief & - \\
        \hline
        Catholic, church, priest, baptized, communion, religion, Catholicism, convert, prayers & practice & inactive \\
        \hline
    \end{tabularx}
    \caption{Practice and belief-related topics from \cite{ifergan}, and a partial list of the words they contain.}
    \label{topics}
\end{table}

\section{Thesaurus Based Evaluation}
\label{sec:list_of_terms}
List of terms from the thesaurus that align with our definitions for religious content:
\begin{itemize}
    \item \textbf{Active Practice} \\
Ritual circumcision · (bio) mohelim · Jewish religious observances · Jewish schools · synagogue attendance · mikva'ot · Yahrzeit · Jewish dietary laws · ghetto Jewish religious observances · camp Jewish religious observances · refugee camp Jewish religious observances · hiding-related Jewish religious observances · Islamic prayers · yizkor · Kaddish · prison Jewish religious observances · Jewish mourning customs · yeshivot · forced labor battalion Jewish religious observances · Hasidic rebbes · (bio) synagogue organizations · (bio) Baalei Keriah · (bio) Baalei Tefillah · (bio) synagogues' sisterhood · (bio) synagogues' men's clubs · deportation Jewish religious observances · Jewish Theological Seminary of America · transfer Jewish religious observances · Beth Jacob schools · b'nai mitzvah · b'nai mitzvah (stills) · Borerim · Jewish religious observances (stills) · Institute of Jewish Studies · shamas · Jewish Institute of Religion · Mitnagdim · observant/practicing · Ramah Camping · Movement · Shema Yisrael.

\item \textbf{Inactive Practice} \\
Christian religious observances · church attendance · religious identity · communions (stills) · confirmations (stills) · Jehovah's Witness missionary activities · Jehovah's Witness religious observances · Mormon missionary activities · Jehovah's Witness religious beliefs · camp Jehovah's Witness religious observances · baptisms · forced labor battalion Jehovah's Witness religious observances · camp Christian religious observances · Eucharist · Islamic religious observances · baptisms (stills) · prison Jehovah's Witness religious observances · ghetto Christian religious observances · Christian religious observances (stills) · confirmations · Islamic dietary laws · Seventh-Day Adventist missionary activities · non-observant/non-practicing · Buddhist religious observances · Buddhist lunar days · karma · Christian missionary activities · Christian prayers.

\item \textbf{Practice - Other} \\
Rabbis.

\item \textbf{Positive Belief} \\
Prayers · Jewish prayers · camp Jewish prayers · Jewish religious beliefs · prison Jewish prayers · forced labor battalion Jewish prayers · camp prayers · ghetto prayers · deportation Jewish prayers · hiding-related prayers · forced march Jewish prayers.

\item \textbf{Negative Belief} \\
Christian religious beliefs · camp Jehovah's Witness prayers · Jehovah's Witness prayers · Islamic identity · Buddhist religious beliefs · Islamic religious beliefs · Armenian Genocide faith issues · Bosnian War and Genocide faith issues · Guatemalan Genocide faith issues · Holocaust faith issues · Rwandan Tutsi Genocide faith issues.
    
\end{itemize}

\section{Evaluation on a Subset of Trajectories}
\label{sec:evaluation_subset}

Besides the evaluation of the full dataset, we randomly sampled 101 testimonies and compared their predicted trajectories in three settings:
prompting GPT-4o and the trained Mistral-7B on religious content predictions, and separately prompting Mistral on the full set of segments. Evaluation results for this subset are provided in Tables \ref{gpt-4o-eval}, \ref{Mistral7B-allcontent} and \ref{Mistral7B-filtered}. We observe that prompting on all segments, as opposed to just religious content, leads to a significant over-prediction rate for each class. Specifically, when using the non-filtered data, the proportion of positive classes does not align with the proportion of religious content in the full dataset, as predicted by a reliable classifier trained on this data.

\begin{table*}[ht]
    \centering
\begin{tabularx}{\textwidth}{l >{\centering\arraybackslash}X >{\centering\arraybackslash}X>{\centering\arraybackslash}X|| >{\centering\arraybackslash}X >{\centering\arraybackslash}X >{\centering\arraybackslash}X >{\centering\arraybackslash}X >{\centering\arraybackslash}X}    \hline
    & \multicolumn{3}{c||}{\textbf{Topic}} & \multicolumn{5}{c}{\textbf{Thesaurus}} \\
    & \multicolumn{1}{c}{\textbf{B}} & \multicolumn{1}{c}{\textbf{P}} & \multicolumn{1}{c||}{\textbf{P\textsuperscript{-}}} & \multicolumn{1}{c}{\textbf{B}} & \multicolumn{1}{c}{\textbf{P}} & \multicolumn{1}{c}{\textbf{P\textsuperscript{+}}} & \multicolumn{1}{c}{\textbf{B\textsuperscript{+}}} & \multicolumn{1}{c}{\textbf{B\textsuperscript{-}}} \\
    \hline
    \hline

Predicted           & \textbf{3.92} & \textbf{32.53} & \textbf{18.46}          & \textbf{6.74} & \textbf{21.87} & \textbf{25.64} & \textbf{2.74} & \textbf{8.06} \\
Original scatter    & 5.60          & 33.33          & \textit{\textbf{17.99}} & 8.22          & 23.39          & 30.81          & 4.80          & 8.92          \\
Edges \& middle     & 5.89          & 33.33          & 19.64          & 8.90          & 23.72          & 26.82          & 5.26          & 9.69          \\
Gaussian            & 6.62          & 41.80          & 19.53          & 8.89          & 29.57          & 34.95          & 3.19          & 9.71          \\
G- Edges \& middle  & 7.07          & 41.30          & 20.47          & 9.65          & 29.05          & 33.70          & 5.17          & 9.54          \\
Equal scatter       & 7.31          & 33.15          & 19.81          & 10.47         & 23.07          & 27.89          & 3.06          & 14.0          \\
\# Reference paths  & 38            & 91             & 26             & 31            & 79             & 76             & 8             & 22            \\
\# predicted paths  & 71            & 88             & 60             & 71            & 88             & 81             & 56            & 35            \\
\# Reference points & 50            & 264            & 40             & 48            & 233            & 206            & 19            & 27            \\
\# predicted points & 368           & 1279           & 157            & 368           & 1279           & 639            & 189           & 80            \\ \hline

\hline

\end{tabularx}
    \caption{Sum of $min\_sum\_dist$ for GPT-4o on a subset of trajectories}
    \label{gpt-4o-eval}
\end{table*}

\begin{table*}[ht]
    \centering
\begin{tabularx}{\textwidth}{l >{\centering\arraybackslash}X >{\centering\arraybackslash}X>{\centering\arraybackslash}X|| >{\centering\arraybackslash}X >{\centering\arraybackslash}X >{\centering\arraybackslash}X >{\centering\arraybackslash}X >{\centering\arraybackslash}X}    \hline
    & \multicolumn{3}{c||}{\textbf{Topic}} & \multicolumn{5}{c}{\textbf{Thesaurus}} \\
    & \multicolumn{1}{c}{\textbf{B}} & \multicolumn{1}{c}{\textbf{P}} & \multicolumn{1}{c||}{\textbf{P\textsuperscript{-}}} & \multicolumn{1}{c}{\textbf{B}} & \multicolumn{1}{c}{\textbf{P}} & \multicolumn{1}{c}{\textbf{P\textsuperscript{+}}} & \multicolumn{1}{c}{\textbf{B\textsuperscript{+}}} & \multicolumn{1}{c}{\textbf{B\textsuperscript{-}}} \\
    \hline
    \hline

Predicted           & \textbf{2.81} & \textbf{0.45} & \textbf{1.87}          & \textbf{0.93} & \textbf{0.34} & \textbf{0.68} & \textbf{2.23} & \textbf{10.35} \\
Original scatter    & 5.27          & 1.22          & 3.38 & 3.71          & 2.33          & 1.84          & 2.36          & 13.00          \\
Edges \& middle     & 5.00          & 2.16          & 2.77                   & 2.53          & 2.65          & 2.32          & 4.67          & 12.31          \\
Gaussian            & 4.13          & 7.07          & 3.03                   & 3.02          & 5.81          & 6.10          & 2.38          & 13.38          \\
G- Edges \& middle  & 5.96          & 7.47          & 4.25                   & 4.53          & 6.07          & 6.56          & 4.71          & 12.44          \\
Equal scatter       & 3.07          & 0.90          & 3.56                   & 2.36          & 0.95          & 1.36          & 1.98          & 16.41          \\
Normal-original  & 3.89          & 3.80          & 2.89                   & 2.68          & 3.43          & 3.95          & 3.89          & 12.81          \\ \hline
\# Reference paths  & 38            & 91            & 26                     & 31            & 79            & 76            & 8             & 22             \\
\# predicted paths  & 93            & 101           & 86                     & 93            & 101           & 101           & 63            & 29             \\
\# Reference points & 50            & 264           & 40                     & 48            & 233           & 206           & 19            & 27             \\
\# predicted points & 507           & 2967          & 328                    & 507           & 2967          & 2385          & 206           & 54             \\ \hline

\hline

\end{tabularx}
    \caption{Sum of $min\_sum\_dist$ of fine-tuned Mistral7B on a subset of trajectories, prompting all content}
    \label{Mistral7B-allcontent}
\end{table*}

\begin{table*}[ht]
    \centering
\begin{tabularx}{\textwidth}{l >{\centering\arraybackslash}X >{\centering\arraybackslash}X>{\centering\arraybackslash}X|| >{\centering\arraybackslash}X >{\centering\arraybackslash}X >{\centering\arraybackslash}X >{\centering\arraybackslash}X >{\centering\arraybackslash}X}    \hline
    & \multicolumn{3}{c||}{\textbf{Topic}} & \multicolumn{5}{c}{\textbf{Thesaurus}} \\
    & \multicolumn{1}{c}{\textbf{B}} & \multicolumn{1}{c}{\textbf{P}} & \multicolumn{1}{c||}{\textbf{P\textsuperscript{-}}} & \multicolumn{1}{c}{\textbf{B}} & \multicolumn{1}{c}{\textbf{P}} & \multicolumn{1}{c}{\textbf{P\textsuperscript{+}}} & \multicolumn{1}{c}{\textbf{B\textsuperscript{+}}} & \multicolumn{1}{c}{\textbf{B\textsuperscript{-}}} \\
    \hline
    \hline

Predicted           & \textbf{3.62} & \textbf{30.09} & \textbf{17.38}          & \textbf{5.68} & \textbf{19.62} & \textbf{15.78} & \textbf{3.66} & \textbf{12.35} \\
Original scatter    & 5.20          & 33.23          & 18.34 & 9.20          & 23.99          & 18.53          & \textbf{3.39} & 15.70          \\
Edges \& middle     & 4.86          & 31.78          & 18.47                   & 7.33          & 22.37          & 18.47          & 6.02          & 14.26          \\
Gaussian            & 5.65          & 36.84          & 19.29                   & 8.40          & 25.58          & 21.22          & 4.14          & 14.83          \\
G- Edges \& middle  & 6.16          & 38.46          & 19.67                   & 9.16          & 27.13          & 23.73          & 6.09          & 14.02          \\
Equal scatter       & 4.04          & 31.45          & 18.93                   & 8.22          & 21.14          & 17.43          & 3.75          & 17.99          \\
Normal-original  & 4.23          & 33.91          & 19.26                   & 8.13          & 23.51          &                & 5.53          & 14.65          \\ \hline
\# Reference paths  & 38            & 91             & 26                      & 31            & 79             & 76             & 8             & 22             \\
\# predicted paths  & 77            & 88             & 69                      & 77            & 88             & 86             & 51            & 26             \\
\# Reference points & 50            & 264            & 40                      & 48            & 233            & 206            & 19            & 27             \\
\# predicted points & 391           & 1816           & 199                     & 391           & 1816           & 1468           & 198           & 46             \\ \hline

\hline

\end{tabularx}
    \caption{Sum of $min\_sum\_dist$ of fine-tuned Mistral7B on a subset of trajectories, prompting filtered content}
    \label{Mistral7B-filtered}
\end{table*}

\section{Structure-Based Taxonomy Analysis}
\label{bins}
\begin{figure*}[ht]
    \centering
    \includegraphics[width=\linewidth]{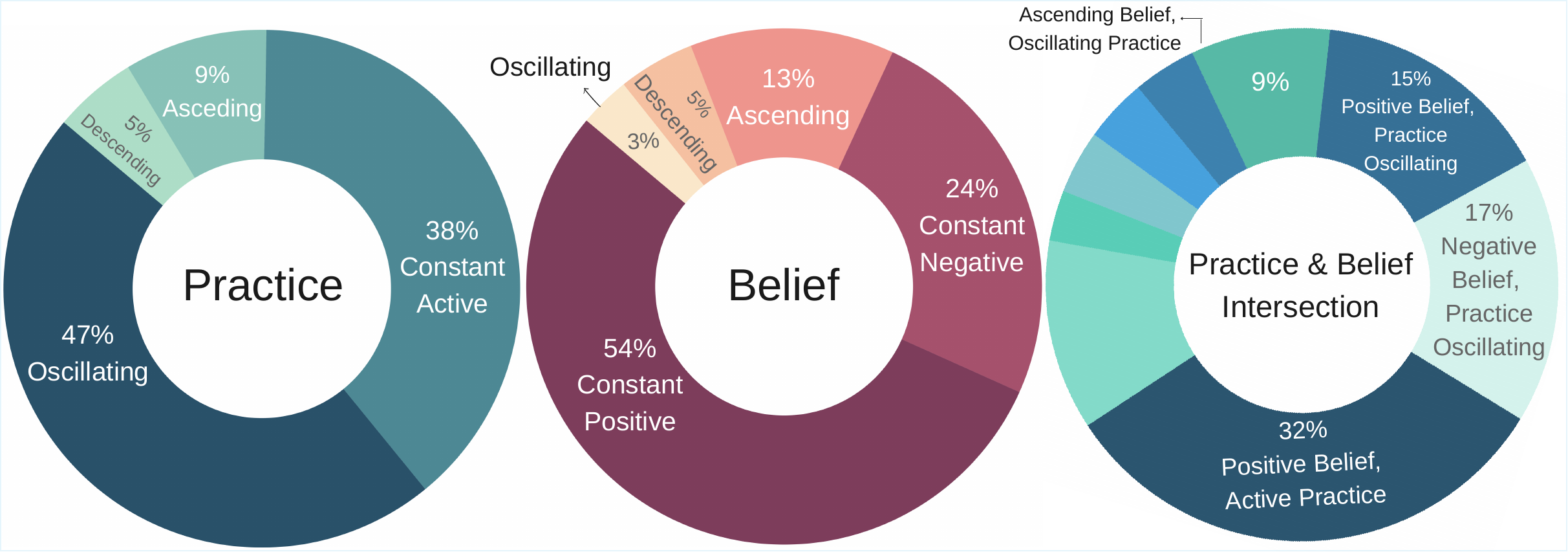}
    \caption{Religious Trajectory structure distributions, for 134 trajectories of length 2-3 for belief and 2-13 for practice.}
    \label{fig:structure_distributions_short}
\end{figure*}

\begin{figure*}[ht]
    \centering
    \includegraphics[width=\linewidth]{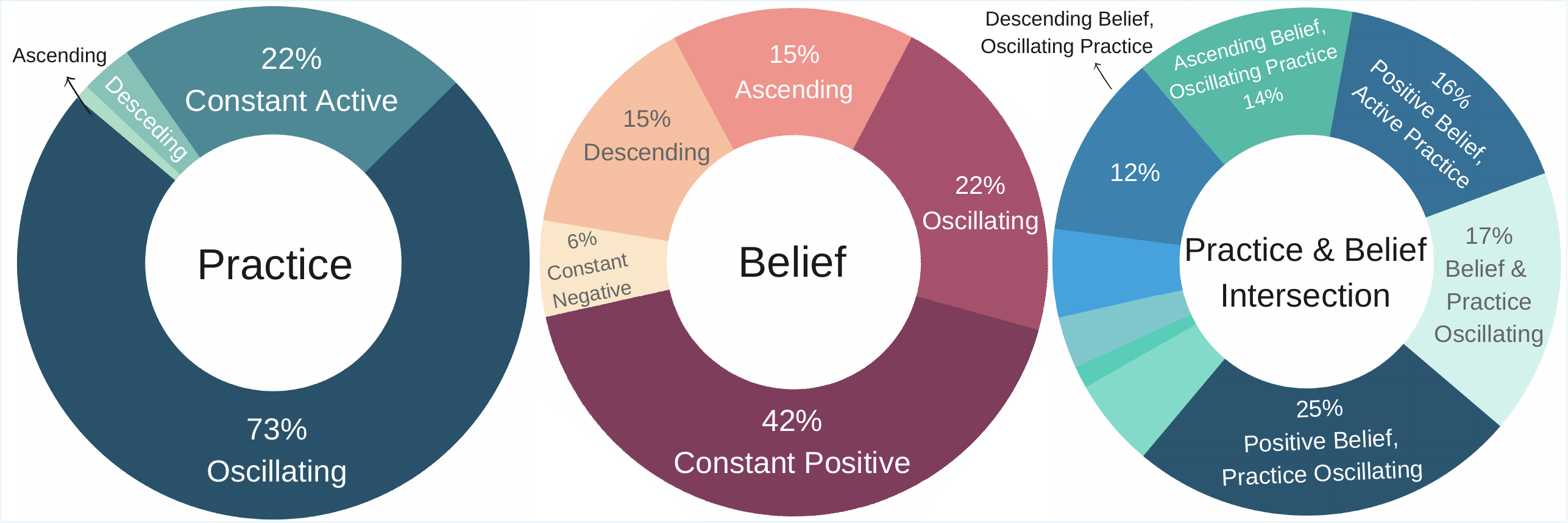}
    \caption{Religious Trajectory structure distributions, for 215 trajectories of length 4-8 for belief and 14-29 for practice.}
    \label{fig:structure_distributions_med}
\end{figure*}

\begin{figure*}[ht]
    \centering
    \includegraphics[width=\linewidth]{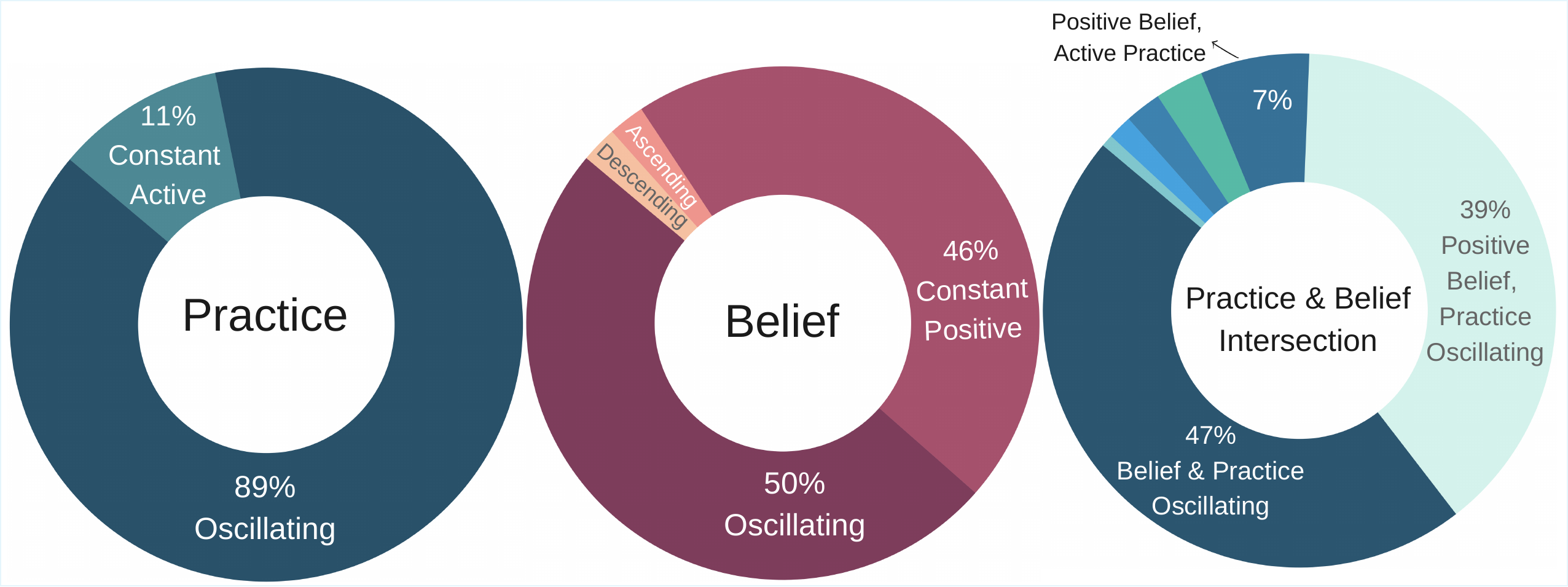}
    \caption{Religious Trajectory structure distributions, for 131 trajectories of length > 8 for belief and > 29 for practice.}
    \label{fig:structure_distributions_long}
\end{figure*}

To examine how density affects the structural distribution of the trajectories, we plot the structure distribution for multiple bins, the sizes of which are inspired by the trajectory length distribution. Results in Figures: \ref{fig:structure_distributions_short}, \ref{fig:structure_distributions_med}, \ref{fig:structure_distributions_long}.


\section{Hierarchical Clustering}
\label{hierarchical_clustering}
\textbf{Belief Clustering Example} 

\begin{figure*}[t]
    \centering
    \begin{minipage}[b]{0.45\textwidth}
        \centering
        \includegraphics[width=\textwidth]{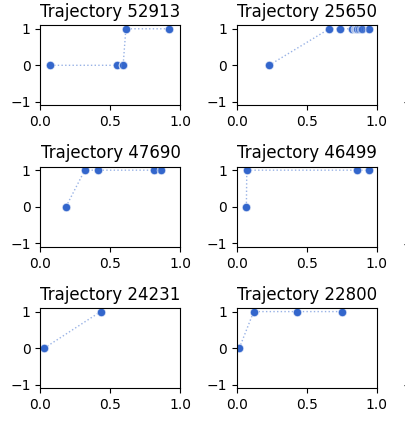}
        \caption*{(a) Other, Ascending}
        \label{fig:sub1b}
    \end{minipage}
    \hspace{0.05\textwidth}
    \begin{minipage}[b]{0.45\textwidth}
        \centering
        \includegraphics[width=\textwidth]{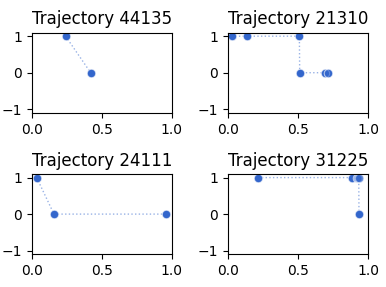}
        \caption*{(b) Positive, Other}
        \label{fig:sub2b}
    \end{minipage}

    \vspace{1em}

    \begin{minipage}[b]{0.45\textwidth}
        \centering
        \includegraphics[width=\textwidth]{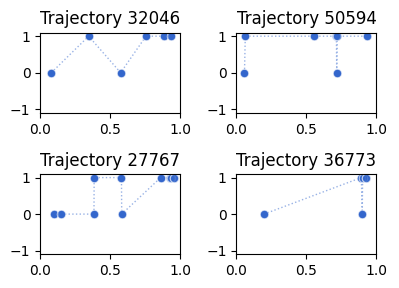}
        \caption*{(c) Other and Positive oscillating}
        \label{fig:sub3b}
    \end{minipage}
    \hspace{0.05\textwidth}
    \begin{minipage}[b]{0.45\textwidth}
        \centering
        \includegraphics[width=\textwidth]{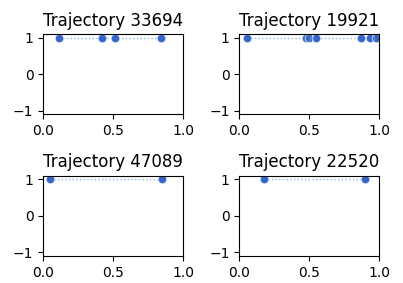}
        \caption*{(d) Positive}
        \label{fig:sub4b}
    \end{minipage}

    \vspace{1em}

    \begin{minipage}[b]{0.45\textwidth}
        \centering
        \includegraphics[width=\textwidth]{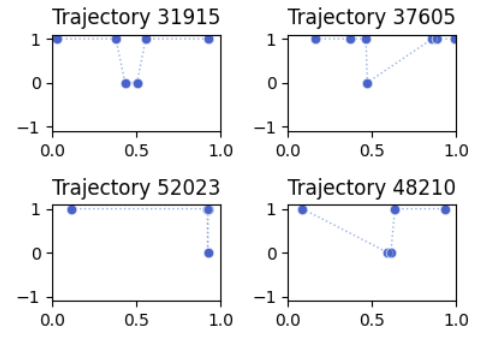}
        \caption*{(e) Positive, Other, Positive}
        \label{fig:sub5b}
    \end{minipage}
    \hspace{0.05\textwidth}
    \begin{minipage}[b]{0.45\textwidth}
        \centering
        \includegraphics[width=\textwidth]{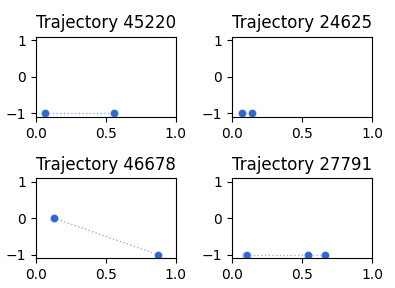}
        \caption*{(f) Sparse Negative}
        \label{fig:sub6b}
    \end{minipage}

    \caption{Belief cluster examples}
    \label{fig:belief_clusters}
\end{figure*}

An example from running HDBSCAN on the belief trajectories: \autoref{fig:belief_clusters}

We first truncated the values of the position points of the trajectories to two decimal points, then calculated their distance matrix using dtw\_ndim.distance\_matrix from the \textit{dtaidistance} Python module, with window=7 and the default values for the rest of the hyperparameters; followed by calling HDBSCAN from the \textit{hdbscan} module, with: min\_cluster\_size=30, min\_samples=1, cluster\_selection\_epsilon=1 and alpha=1. 

\paragraph{Practice Clustering Example} 

\begin{figure*}[t]
    \centering
    \begin{minipage}[b]{0.45\textwidth}
        \centering
        \includegraphics[width=\textwidth]{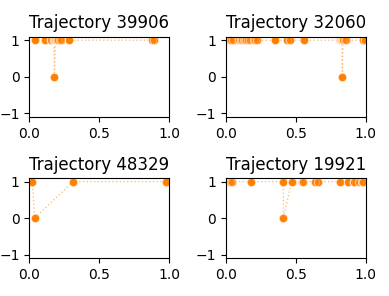}
        \caption*{(a) Active, Other, Active}
        \label{fig:sub1a}
    \end{minipage}
    \hspace{0.05\textwidth}
    \begin{minipage}[b]{0.45\textwidth}
        \centering
        \includegraphics[width=\textwidth]{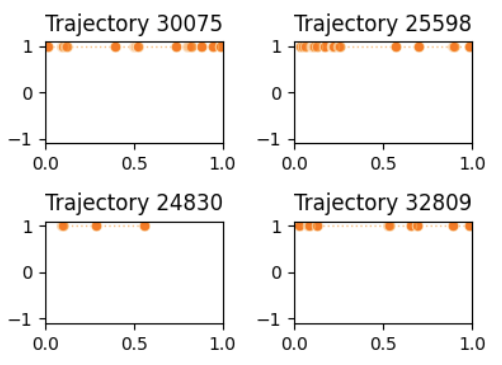}
        \caption*{(b) Active}
        \label{fig:sub2a}
    \end{minipage}

    \vspace{1em}

    \begin{minipage}[b]{0.45\textwidth}
        \centering
        \includegraphics[width=\textwidth]{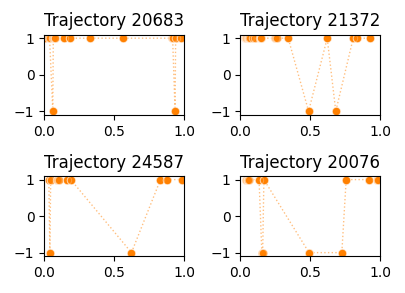}
        \caption*{(c) Oscillating}
        \label{fig:sub3}
    \end{minipage}
    \hspace{0.05\textwidth}
    \begin{minipage}[b]{0.45\textwidth}
        \centering
        \includegraphics[width=\textwidth]{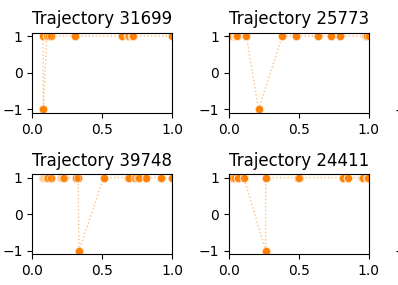}
        \caption*{(d) Active, Inactive, Active}
        \label{fig:sub4}
    \end{minipage}

    \caption{Overall figure caption}
    \label{fig:prac_clusters}
\end{figure*}

An example from running HDBSCAN on the practice trajectories: \autoref{fig:prac_clusters}\\
Hyper-parameters: window=6, min\_cluster\_size=30, min\_samples=1, cluster\_selection\_epsilon=1 and alpha=0.95.

\section{Classification Performance Analysis}
\label{conf_matrices}

\begin{figure*}[h!]
  \centering
  \includegraphics[width=0.46\textwidth]{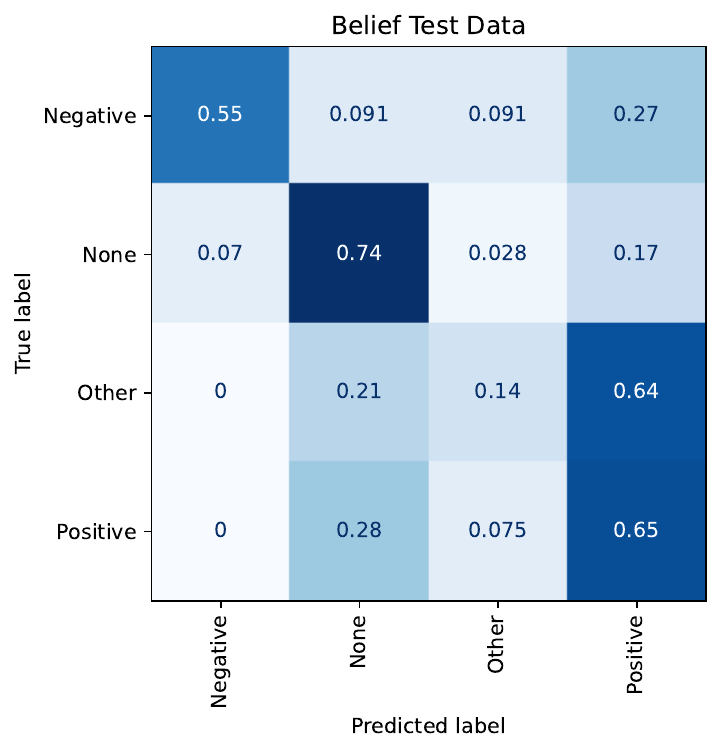}
  \hspace{0.05\textwidth} 
  \includegraphics[width=0.46\textwidth]{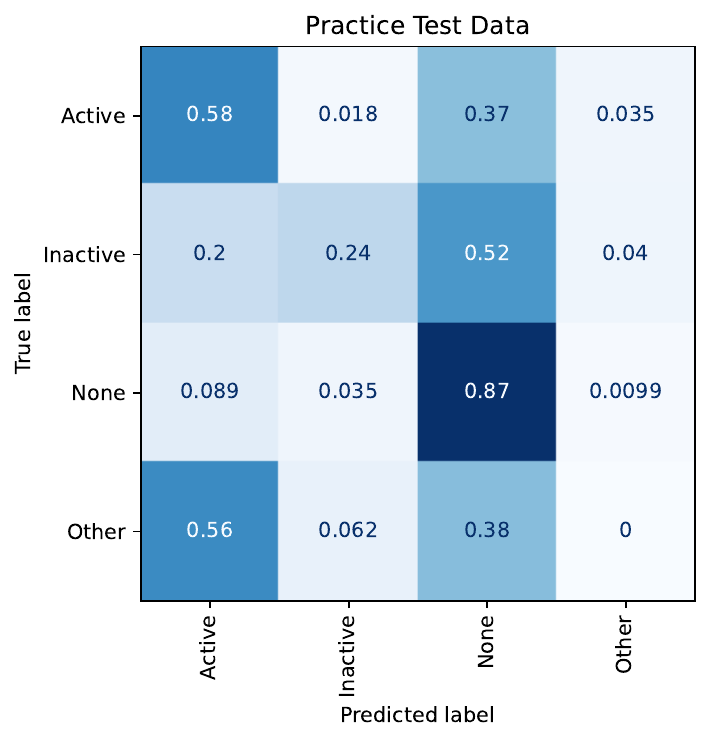}
  \caption{Practice and Belief classification confusion matrices}
  \label{fig:conf_matrices}
\end{figure*}

The confusion matrices in \autoref{fig:conf_matrices} summarize the classification performance.

\end{document}